\documentclass[10pt,twocolumn,letterpaper]{article}

\usepackage[pagenumbers]{cvpr} % To force page numbers, e.g. for an arXiv version

\usepackage{graphicx}
\usepackage{amsmath}
\usepackage{amssymb}
\usepackage{booktabs}

\usepackage{algorithm}
\usepackage{algpseudocode}

\usepackage{tabularx}
\usepackage{multirow}
\usepackage{boldline}
\usepackage[pagebackref,breaklinks,colorlinks]{hyperref}

\usepackage{array}
    \newcolumntype{P}[1]{>{\centering\arraybackslash}p{#1}}
    \newcolumntype{M}[1]{>{\centering\arraybackslash}m{#1}}

\usepackage{hyperref}
\hypersetup{
    colorlinks=true,
    linkcolor=blue,
    filecolor=magenta,     
    urlcolor=blue,
}
\usepackage{url}
\usepackage{graphicx}
\usepackage{color, colortbl}
\definecolor{LightCyan}{rgb}{0.88,1,1}
\definecolor{LightGreen}{rgb}{.886,  .937,  .855}
\definecolor{LightGrey}{rgb}{.86, .86, .86}

\usepackage{amsmath}

\usepackage[capitalize]{cleveref}
\crefname{section}{Sec.}{Secs.}
\Crefname{section}{Section}{Sections}
\Crefname{table}{Table}{Tables}
\crefname{table}{Tab.}{Tabs.}

\def\cvprPaperID{2113} % *** Enter the CVPR Paper ID here
\def\confName{CVPR}
\def\confYear{2022}

\begin{document}

% \title{Efficient Spatial Separable Transformer for High Resolution Semantic Segmentation}
\title{Dual-Flattening Transformers through Decomposed Row and Column Queries for Semantic Segmentation}
\author{
Ying Wang$^{1}$
\and
Chiuman Ho$^{1}$
\and
Wenju Xu$^{1,2}$
\and
Ziwei Xuan\thanks{Work conducted during an internship at OPPO US Research Center.}$\;^{3}$
\and
Xudong Liu$^{1}$
\and 
Guo-Jun Qi\thanks{Corresponding author: Guo-Jun Qi, Email: guojunq@gmail.com.}\;$^{1}$ \\
$^{1}$ OPPO US Research Center\quad
$^{2}$ InnoPeak Technology, Inc.\quad
$^{3}$Texas A\&M University\\
{\tt\small \{ying.wang, chiuman, wenju.xu, xudong.liu\}@oppo.com, xuan64@tamu.edu, guojunq@gmail.com}
}
% \institute{Oppo US Research\\
% \email{\{ying.wang, chiuman, wenju.xu, ziwei.xuan, xudong.liu, guojun.qi\}@innopeaktech.com} \and
% Texas A\& M University\\
% \email{ziwei.xuan@innopeaktech.com}}
% \author{Ying Wang\\
% Oppo US Research\\
% {\tt\small ying.wang@innopeaktech.com}
% \and
% Chiuman Ho\\
% Oppo US Research\\
\maketitle

%%%%%%%%% ABSTRACT
\begin{abstract}
  It is critical to obtain high resolution features with long range dependency for dense prediction tasks such as semantic segmentation. To generate high-resolution output of size $H\times W$ from a low-resolution feature map of size $h\times w$ ($hw\ll HW$), a naive dense transformer incurs an intractable complexity of $\mathcal{O}(hwHW)$, limiting its application on high-resolution dense prediction. We propose a Dual-Flattening Transformer (DFlatFormer) to enable high-resolution output by reducing complexity to $\mathcal{O}(hw(H+W))$ that is multiple orders of magnitude smaller than the naive dense transformer. Decomposed queries are presented to retrieve row and column attentions tractably through separate transformers, and their outputs are combined to form a dense feature map at high resolution.  To this end, the input sequence fed from an encoder is row-wise and column-wise flattened to align with decomposed queries by preserving their row and column structures, respectively. Row and column transformers also interact with each other to capture their mutual attentions with the spatial crossings between rows and columns. We also propose to perform attentions through efficient grouping and pooling to further reduce the model complexity. Extensive experiments on ADE20K and Cityscapes datasets demonstrate the superiority of the proposed dual-flattening transformer architecture with higher mIoUs.
  %at a comparable model size and reduced computational overhead.
\end{abstract}

%%%%%%%%% BODY TEXT
\section{Introduction}
Obtaining high-resolution  features is important in many computer vision tasks, especially for dense prediction problems such as semantic segmentation, object detection and pose estimation. Typical approaches employ convolutional encoder-decoder architectures where the encoder outputs low-resolution features and decoder upsamples features with simple  filters such as bilinear interpolation. Bilinear upsampling has limited capacity in obtaining high-resolution features, as it only conducts linear interpolation between neighboring pixels without considering nonlinear dependencies in global contexts. Various approaches have been proposed to improve the high-resolution feature quality, such as PointRend~\cite{kirillov2020pointrend} and DUpsample~\cite{tian2019decoders}. PointRend~\cite{kirillov2020pointrend} carefully selects uncertain points in the downsampled feature space and refines them by incorporating low-level features. DUpsample~\cite{tian2019decoders} adopts a data-dependent upsampling strategy to recover segmentation from the coarse prediction. However, these approaches lack the ability in capturing long-range dependency for  fine-grained details. Diverse non-local or self-attention based schemes have been proposed to enhance the output features~\cite{cao2019gcnet, hu2019acnet, huang2019ccnet, wang2020axial}, but mostly in the downsampled feature space. They still rely on bilinear upsampling procedure to obtain high-resolution features which tends to lose global information.
 
%  For dense prediction, typical approaches output downsampled prediction with CNNs,  and upsample output features with bilinear interpolation.

Recently, transformers have drawn tremendous interests, due to its great success in capturing long-range dependency.  As the first standard-alone transformer for computer vision,  Vision Transformer(ViT)~\cite{dosovitskiy2020image} shows impressive result on image classification with patch-based self-attentions. However there is still a large room to improve when applying to dense prediction tasks due to its low-resolution feature map caused by non-overlapping patches and intractable computing costs. The complexity of a naive dense transformer scales rapidly w.r.t the high-resolution output, limiting its application to dense prediction problems. Multi-scale ViTs~\cite{liu2021swin, wang2021pyramid, xie2021segformer, wu2021p2t} have been presented to achieve hierarchical features with different resolutions and have boosted the performance of many dense prediction tasks.  However, upper-level features with low spatial resolution still rely on bilinear upsampling to recover the full-resolution features.   
%Most of existing approaches output downsampled features and utilize bilinear upsampling to recover to the original resolution. 
The naive bilinear upsampling is inherently weak since it is intrinsically linear and local in recovering fine-grained details by linearly interpolating from local neighbors.   

%Different from self-attention based transformers, DETR~\cite{carion2020end} directly models the object detection task as set matching problem and achieves comparable performance with fully optimized convlutional neural network (CNN) based detection frameworks.
%It retrieves object information with learnable queries. It builds a transformer decoder on top of a convolutional encoder, where the encoder output features are considered as keys and multi-head attention is conducted between learnable queries and keys to retrieve object information. The architecture shows a strong modeling capacity  under properly designed signals. However, it also suffers from heavy computational cost for high-resolution feature maps. A series of works have been presented to address the complexity issue.  Deformable DETR~\cite{zhu2020deformable} employs deformable attention on a sparse set of feature representations, which significantly reduces the memory and computation cost. However, the learned sparse pattern leads to irregular access of memory which may restrict its application. 

Several efficient attention designs  can reduce the model complexity, such as  Axial-attention~\cite{wang2020axial}, Criss-Cross attention~\cite{huang2019ccnet} and LongFormer~\cite{beltagy2020longformer}. However, they mainly focus on feature enhancement in the downsampled space without recovering the high-resolution features or recovering the fine-grained details by modeling the nonlinear dependency on a more global scale from non-local neighbors. Therefore, in this paper, we strive to develop a transformer architecture that is not only efficient to recover full-resolution features but also able to recover the fine-grained details by exploring full contexts nonlinearly and globally.

For this purpose, to reduce the complexity w.r.t. to the high-resolution size, we propose a Dual-Flattening transFormer (DFlatFormer) with decomposed row and column queries to capture global attentions.  The key idea is to decompose the output feature space into row and column ones, and utilize the corresponding queries to output row and column representations separately. The final feature map can be obtained by integrating the two outputs. Let the size of a low-resolution input be $h\times w$, and the target high-resolution output  size be $H\times W$. Our model only incurs a complexity of $\mathcal{O}\big(hw(H+W)\big)$, while the naive dense transformer has a much larger complexity of $\mathcal{O}(hwHW)$ as $h\ll H$ and $w\ll W$.
%For example, suppose that the size of a high-resolution  output feature map is $H\times W=512\times 512$ and the size of a low-resolution input feature map is  $h\times w=128\times 128$. The naive dense transformer has a complexity  $\frac{HW}{H+W}=256$ times larger than ours, which is intractable in practice. 
Our motivation is partially analogy to depth-wise separable convolution~\cite{chollet2017xception} which decomposes regular convolution into depth-wise and point-wise ones for reducing the complexity, whereas ours decompose a transformer into rows and columns before they interact with each other and are eventually re-coupled to form a dense feature map. 
% To facilitate row/column queries to retrieve information from more spatially related positions, the encoder output feature map is row/column-wise flattened to serve as corresponding keys. The  row-flattened keys enable gathering more closely row-related semantic information. 
% \textcolor{blue}{Inspired by the powerful relation modeling capability of query based framework such as  DETR~\cite{carion2020end}, we adopt a similar transformer encoder-decoder architecture and employ cross-attention and row/column queries to retrieve long-range information.}
% The flattening structure also makes it easier to obtain global keys by pooling over more semantic related regions.

To accommodate with the decomposed row and column queries, we propose to flatten an input feature map row-wise and column-wise into two separate sequences, so that successive rows and columns are put together to preserve their row and column structures, respectively. This will enable row and column queries to retrieve their immediate contexts from the neighboring rows and columns. In this way, the dual-flattened sequences will ideally align with the query sequences of rows and columns, resulting in a dual-flattening transformer architecture presented in this paper. 

With these dual-flattened sequences as input, the DFlatFormer employs a row and a column transformer, each generating  row and column outputs at a higher resolution, respectively. Each transformer is composed of several layers of multi-head attentions  between the decomposed queries and dual-flattened inputs followed by feed-forward networks (FFNs). Moreover, to refine the spatial details, we introduce row-column interactive attentions to reflect the crossing points between rows and columns. In this way, the sequences of row and column outputs will fully interact to share their learned representations in their vertical and horizontal contexts when they spatially cross each other. To further reduce the complexity, we apply grouping and pooling to query sequences and/or flattened inputs, enabling more efficient computing of attentions.

%add a key selection module right after row/column-wise flattening to reduce the number of keys retrievable by each query. Local and global keys are selected through grouping and pooling to  retrieve both short and long range contexts.

% Our motivations are twofolds. First, feature representations in images are typically redundant, taking all the pixels as keys  incurs heavy computation and slow convergence;  Second, pooled feature representations provide more abstract and global context. Utilizing both fine-grained local features and globally  pooled ones potentially provide supplementary information and may provide more efficient learning.

% Row and column queries are learned embedding. The initialization is taken as the interpolation from key positional code for alignment. 

% To retrieve global contextual information for each row in the target feature map, the row query has to associate with more related keys. 

Our major contributions are summarized below.
\begin{itemize}
    \item We propose an efficient dual-flattening transformer architecture to obtain high-resolution feature map, with a complexity of $\mathcal{O}(hw(H+W))$ where $h\times w$ and $H\times W$ are the input and output feature map sizes, respectively. The proposed architecture can serve as a flexible plug-in module into any CNN or transformer-based encoders to obtain high-resolution dense predictions.
    \item For this purpose, we propose to flatten input sequences row-wise and column-wise before they are fed into multi-head attentions along with decomposed queries. Interactive attentions are also presented to share vertical and horizontal contexts between rows and columns.   
    \item We also propose efficient attentions via grouping and pooling in the feature space, which further reduces complexity to $\mathcal{O}\big((\beta_g+\beta_p)hw(H+W)\big)$, with $\beta_g$ and $\beta_p$ the fractions of grouped and pooled features, respectively. 
    \item Experimental results  demonstrate the superior performance of DFlatFormer as a universal plug-in into various CNN and transformer backbones for semantic segmentation  on multiple datasets.
    
\end{itemize}

\section{Related work}
%\subsection{Efficient self-attention}

% \noindent {\bf Vision Transformers.} 

\noindent {\bf Self-attention.}
For CNN based architecture, numerous efforts haven been made on increasing the receptive field, through dilated convolutions or introducing self-attention modules~\cite{vaswani2017attention, fu2019dual,cao2019gcnet, hu2019acnet, huang2019ccnet}.  DANet~\cite{fu2019dual} employs both spatial and channel attention modules to model the semantic inter dependencies for segmentation. Traditional dense self-attention imposes heavy computation and memory cost. Various works haven been proposed with improved efficiency~\cite{wang2020axial,huang2019ccnet}.  Axial self-attention~\cite{wang2020axial} applies local window along horizontal and vertical axis sequentially to achieve global attention. CCNet~\cite{huang2019ccnet} adopts self-attention on its criss-cross path in a recursive manner. These methods have shown impressive capacity in modeling long-range contextual information at a much smaller memory size and complexity.

% \noindent {\bf Upsampling and feature enhancement.}

%\subsection{Efficient transformer}
\noindent {\bf Transformers for vision.} Transformers have been known for their Superior capability in modeling long-range dependency. As a first pure transformer for vision, Vision Transformer(ViT)~\cite{dosovitskiy2020image} has achieved tremendous success in image classification. Rapid and impressive progresses have been made for dense prediction~\cite{liu2021swin,wang2021pyramid,zheng2021rethinking,wu2021p2t,wang2021max,wang2021crossformer,dong2021cswin,xie2021segformer,chu2021twins,yang2021focal}. DETR~\cite{carion2020end} formulates the object detection into set matching problem and  proposes a learnable query based framework to directly retrieve object information. MaX-DeepLab~\cite{wang2021max} employs a  dual-path transformer for panoptic
segmentation which contains a global
memory path in addition to a CNN path. Swin Transformer~\cite{liu2021swin} employs self-attention within local windows but effectively captures long-range dependency through cyclic window shift. Pyramid vision transformer (PVT)~\cite{wang2021pyramid} exploits hierarchical transformer structure to obtain multi-level features which greatly benefits dense prediction. SETR~\cite{zheng2021rethinking} deploys a transformer along with a simple decoder from sequence to sequence perspective.  Recently SegFormer~\cite{xie2021segformer} presented a positional-encoding-free  hierarchically structured Transformer encoder and a lightweight MLP decoder which achieves superior performance with improved computation efficiency. There is a resurgence of efficient transformer design from various aspects including attention pattern, memory, and low-rank methods ~\cite{wang2020linformer, child2019generating,kitaev2019reformer}.

\noindent {\bf Semantic segmentation.}
Since the seminal work of fully convolutional network (FCN)~\cite{long2015fully} leading to significant progress in semantic segmentation, a series of model architectures have  been proposed to further improve performance. Some works target at improving the receptive field of CNNs through dilated convolution~\cite{chen2018encoder}. 
Other works improve fine-grained contexts through hierarchical structures~\cite{lin2017feature,tao2020hierarchical}, exploring object contextual correlation~\cite{yuan2020object}, and feature refinement~\cite{kirillov2020pointrend,tian2019decoders}. Besides CNN-based models, Transformers~\cite{dosovitskiy2020image,liu2021swin,zheng2021rethinking,kitaev2019reformer} have shown their capability in capturing global contexts which are important for segmentation.

\section{The proposed model: DFlatFormer}
In this section, we present in details the proposed Dual-Flattening transFormer (DFlatFormer). Consider a general semantic segmentation framework that consists of an encoder to extract features and a decoder for pixel-wise classification. The encoder typically gives a low-resolution feature map, and the decoder aims to output a dense feature map to enable high-resolution segmentation.

Suppose the encoder outputs a feature map $S_o \in \mathcal{R}^{h\times w \times d}$, and the decoder seeks to output a feature map $S\in \mathcal{R}^{H\times W \times d}$ at a target resolution, where $h(H)$, $w(W)$ and $d$ are the height, width and channel dimension of encoder (decoder) feature map, respectively. Normally $h\ll H$ and $w\ll W$. Our goal is to obtain  a high-resolution feature map $S$ from the encoder output $S_o$ through the DFlatFormer.
% Denote the ratio of target feature map size w.r.t encoder one as $\alpha$, i.e., $H=\alpha H_o$ and $W=\alpha W_o$. The ratio $\alpha>0$ can be any positive real number, indicating the capability of both upsampling and downsampling of our scheme.  Typically $\alpha>1$,  the output of DFlatFormer can be considered as an up-sampled version of the encoder output, packed with rich global contextual information. 

\begin{figure*}[ht!]
    \centering
    \includegraphics[width=\textwidth]{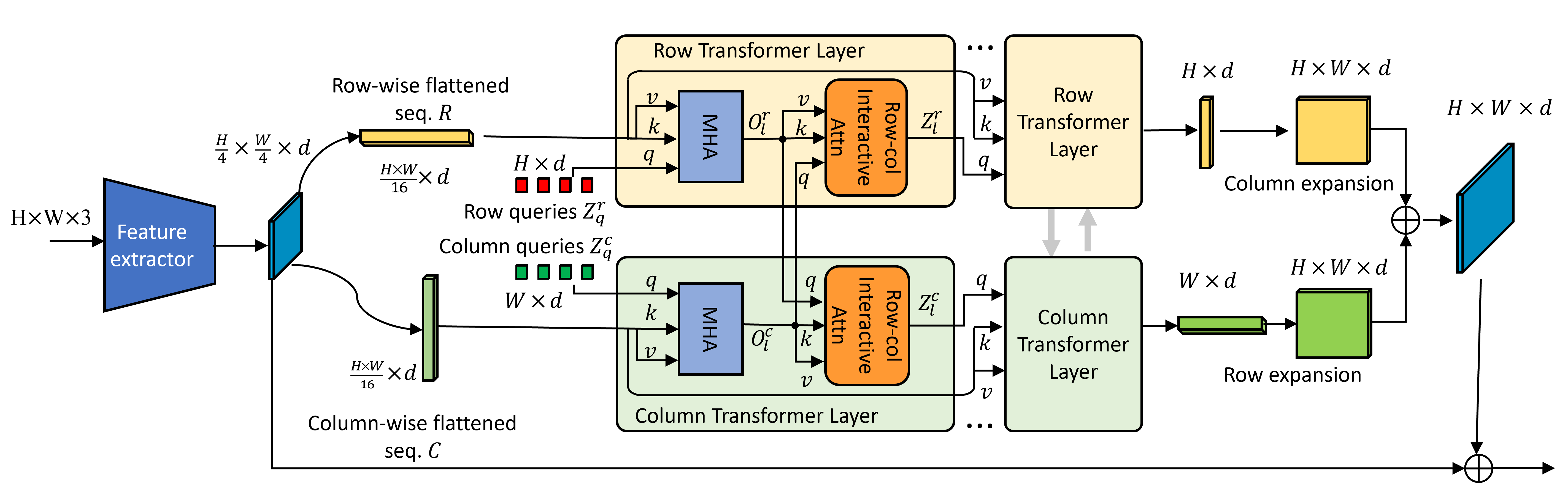}
    \caption{Illustration of DFlatFormer framework. Residual connections and normalization are omitted in the figure for better exposition to avoid unnecessary cluttering.}
    \label{fig:dflatformer_framework}
\end{figure*}

\subsection{The overall framework}

Naively, a dense transformer needs a full-size query sequence to embed a flattened sequence of low-resolution input to a high-resolution output. This is intractable as the full size sequence of $H\times W$ queries would result in demanding memory and computational overheads.

On the contrary, in the proposed DFlatFormer, we decompose full-size queries in a naive dense transformer into $H$ row and $W$ column ones, and use them to embed rows and columns separately through multi-head and interactive attentions  (see Sec.~\ref{sec:mha}). The input sequence from the encoder will be flattened row-wise and column-wise to spatially align with the sequences of decomposed queries (cf. Sec.~\ref{sec:dual_flatten_keys}) before it is fed into row and column transformers separately. 
These embedded row and column representations will further interact with each other to aggregate vertical and horizontal contexts at the row-column crossing points, before they are eventually expanded and combined to generate a dense feature map of high resolution.  

Fig.~\ref{fig:dflatformer_framework} summarizes the overall pipeline. Specially, the DFlatFormer consists of a row transformer and a column transformer. Given the low-resolution output $S_o$ from an encoder, it is flattened row-wise and column-wise into two separate sequences $R$ and $C$ with the respective positional encodings (cf. Sec.~\ref{sec:dual_flatten_keys}). The flattened row and column sequences together with the decomposed sequences of learnable row $Z_q^r$ and column $Z_q^c$ queries are fed through row and column transformers to output corresponding embeddings. The resultant row/column embeddings for each layer are further refined by interacting with their column/row counterpart through an interactive attention module (see Sec.~\ref{sec:mha}).  Suppose the row (column) transformer consists of $L$ layers, the resultant last-layer embeddings of rows $Z^r_{L}$  and columns $Z^c_{L}$ are finally expanded and combined to output a full-resolution feature maps $S$ (see Sec.~\ref{sec:mha}).  
%Below we will present each module in detail. 

Let us take a closer look at each layer of DFlatFormer. Taking the row transformer as an example, a layer  is composed of a multi-head attention (MHA) module and a row-column interactive attention module.
%the process to obtain a high-resolution target feature map is as follows. 
Its MHA module takes as inputs the original row-flattened sequence $R$, the row query sequence $Z^r_q$ and the last layer output of row embeddings $Z^r_{l-1}$, and it outputs $O^r_{l}$ after a multi-layer perceptron (MLP) of Feed-Forward Network (FFN).  Then a row-column interactive attention module follows by coupling intermediate embeddings of $O^r_l$ and $O^c_l$ output from the row and column MHAs. This module allows a row representation to aggregate all column embeddings in vertical contexts as they cross the row.  It outputs the row representation of $Z^r_l$ for layer $l$, which will be fed into the next layer for further modeling. The details will be presented in Sec.~\ref{sec:mha}.

To further improve the computational efficiency, we propose a grouping and pooling module by reducing the number of row/column flattened tokens fed into each layer. The details are explained in Sec.~\ref{sec:key_selection}. DFlatFormer can serve as a plug-in module to be connected to any CNN or transformer based encoder and generate high-resolution output.

\subsection{Dual flattening and positional encodings}\label{sec:dual_flatten_keys}

%With decomposed row and column queries, 
First, we flatten the encoder output $S_o$ row-wise and column-wise to align with the row and column queries separately. This preserves the row and column structures by putting entire rows and columns together in the flattened sequences, respectively. Meanwhile, row-wise and column-wise positional encodings will also be applied to the dual-flatten sequences (cf. Sec.~\ref{sec:pos_coding}). This will enable the row and column queries to apply the multi-head attentions to the respective flattened sequences.

%serve as keys in the row/column direction. Taking row Transformer, $S$ is row-wise flattened to serve as keys. For keys in each row, we utilize the same positional code within the row. The decomposed row and column queries are set as learnable embeddings.

\subsubsection{Row-wise and column-wise positional encodings}\label{sec:pos_coding}

Typically, the target feature map  has much larger size than the encoder output. To align the query sequence with the row/column flattened key/value sequence, we need a row/column-wise positional encodings as well.  

\begin{figure}[ht!]
    \centering
    \includegraphics[width=\columnwidth]{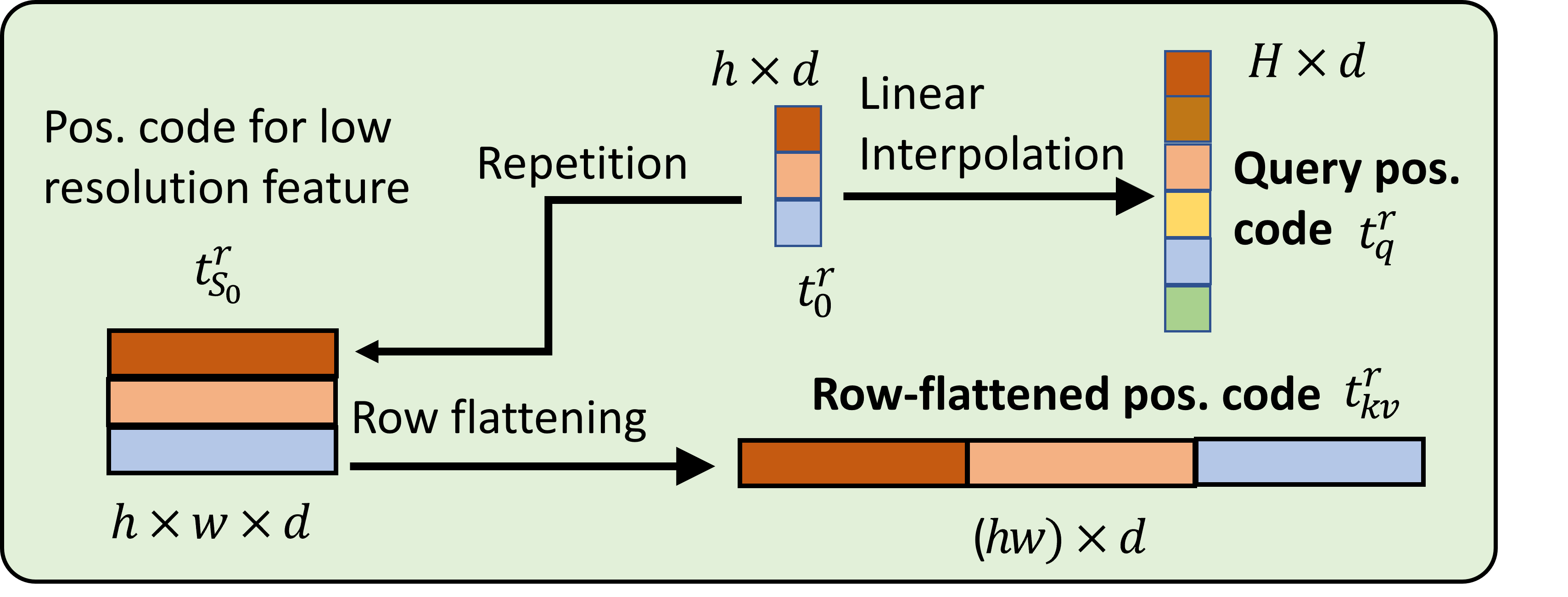}
    \caption{Positional coding of row queries and row-flattened sequence.}
    \label{fig:pos_interp}
\end{figure}

Formally, taking the row part for example, we have a sequence of $H$ row queries. Meanwhile, the row sequence only has $h$ rows before it is row-wise flattened to $R\in\mathcal R^{hw\times d}$.  As shown in Fig.~\ref{fig:pos_interp}, we start with an initial 1D positional encoding $t^r_0\in\mathcal R^{h\times d}$ through sinusoidal function following \cite{vaswani2017attention}. This encoding can be upsampled by linear interpolation to  $t^r_q\in\mathcal R^{H\times d}$, which aligns with the size of query sequence and thus can be used to initialize the positional encoding of queries directly. Meanwhile, $t^r_0$ can also be column-wise replicated to $t^r_{S_0}\in\mathcal R^{h\times w\times d}$. The resultant encoding aligns with the size of input feature map $S_o$, and thus can be row-wise flattened to $t^r_{kv}\in\mathcal R^{hw\times d}$ and used as the positional encoding of the row-wise flattened $R$.  This is a row-wise positional encoding since each row has the same code, thereby aligning with the row-wise flattening. The similar column-wise positional encoding can be applied to the corresponding column-wise flattening.

While the above depicts the mechanism of aligning absolute positional encodings with row/column-wise flattening, it is not hard to apply the similar idea of row/column-wise expansion to extend relative positional encodings \cite{shaw2018self, bello2019attention} as well.  Interested readers can refer to our source code \footnote{It will be available upon the acceptance of this submission.} for more details.

% \noindent {\bf Relative positional coding.}
% Relative positional coding is shown to have advantage over its absolute counterpart in some cases, especially when the input size is different. Relative coding can adapt to input feature map with arbitrary size.  For row Transformer, let $B^r\in \mathcal{R}^{H\times W}$ be the relative position bias. $B^r$ is interpolated from a learnable small set ${B}_{r0}\in \mathcal{R}^{(2h-1)\times (2w-1)}$.  If the query positional code is learnable,  the interpolated values are taken as initialization of $B^r$. The attention is given as
% % {\small
% % \begin{align*}
% %     \texttt{ATTN}(z_q, z_k, z_v)
% %     =\texttt{SoftMax}\big(\frac{(z_qW_q)( z_kW_k)^T}{\sqrt{d}}+B\big)z_vW_v.
% % \end{align*}
% % }

% {\small
% \begin{align*}
%     \texttt{ATTN}(Q^r,K^r,V^r)
%     =\texttt{SoftMax}\big(\frac{Q^r{K^r}^T}{\sqrt{d}}+B^r\big)V^r.
% \end{align*}
% }

% \noindent {\bf Learnable positional embedding.}
% The query positional code is set to learnable embedding, with initialization from bilinearly interpolation from key positional code. The key positional code can be either absolute or relative ones.

% \noindent{\bf Learnable window filter.}
% Besides learnable positional embedding for queries, an alternative is to use learnable window filtering to output the positional code for queries.  It reduces the number of learnable parameters compared with direct embedding. The window filter is of size $n_o\times m_o$ and stride size $s_o$, where $s_o=\frac{W_o}{\alpha}$.  

\subsection{Multi-head and row-column interactive attentions} \label{sec:mha}
% The standard DETR decoder contains two types of attention modules: self-attention and cross-attention. Self-attention enables interaction between object queries to capture relation with each other, while cross-attention captures the relation between object queries with encoder output features. 
% \textcolor{blue}{ We utilize cross-attention as the row/column queries are explicitly designed to learn the corresponding row/column information. } 
{\noindent\bf Multi-head attention.}
First we embed row and column queries separately through multi-head attention with dual-flattened sequences. 
%we adopt multi-head attention (MHA)  to capture long-range dependency.  
%Specifically, suppose there are $n_h$ heads for an MHA and the  embedding dimension is $d_m$ for each head. 
Let $Z_{l-1}^r (Z_{l-1}^c)$ be the row (column) output sequence for layer $l-1$ that will be fed into the next layer as input. The first layer input $Z_0^r (Z_0^c)$ is set to 0. 
%Let $R(C)$ be the row(column)-flattened sequence from $S_o$. 
% \textcolor{blue}{At each layer $l$, the query, key and value embeddings are taken as inputs to MHA, computed as: $Q^r_l=Z^r_{q,l-1}W_q$, $K_l^r=RW_k$ and $V_l^r=RW_v$, where $W_q$, $W_{k}$ and $W_v$ are the linear projection matrices. } 
Then, taking a row transformer for example, a single-head attention is formulated as
{\small
\begin{align} \label{eq:attn}
    &\texttt{ATTN}(Z_{l-1}^r,R,Z_q^r)\\\nonumber
    =&Z_{l-1}^r+\texttt{SoftMax}\big(\frac{(Z_{l-1}^r+Z_q^r) W_q \big((R+t_{kv}^r)W_k\big)^T}{\sqrt{d_m}}\big)RW_v,
\end{align}
}
where $W_q$, $W_{k}$ and $W_v$ are the linear projection matrices, and $d_m$ is the embedding dimension for each head. $Z_q^r$ is the learnable sequence of row queries and $t_{kv}^r$ is the  positional encodings defined in Sec.~\ref{sec:pos_coding}.  Then, the multi-head attention is obtained by putting together single-head outputs
{\small
\begin{align*}
    \tilde{O}_l^r={\big[\texttt{ATTN}_0(Z_{l-1}^r,R,Z_q^r); \cdots;\texttt{ATTN}_{n_h-1}(Z_{l-1}^r,R,Z_q^r) \big]}W^O,
\end{align*}
}
where $W^O$ is a linear projection matrix. 
% ${\texttt{ATTN}(z^r_{q,l,m}W_{q,l,m}^r,z^r_{k,l,m}W_{k,l,m}^r,z^r_{v,l,m}W_{v,l,m}^r)}$
% \begin{align}
%     \tilde{z}_l^r=&\texttt{MHA}(z^r_{q,l}W_{q,l}^r,z^r_{k,l}W_{k,l}^r,z^r_{v,l}W_{v,l}^r)\\ \nonumber
%     =&\texttt{Concat}{\big[{\texttt{ATTN}(z^r_{q,l,m}W_{q,l,m}^r,z^r_{k,l,m}W_{k,l,m}^r,z^r_{v,l,m}W_{v,l,m}^r)}\big]}_{m=0}^{n_h-1}W^O,
% \end{align}
% where $n_h$ is the number of heads, and $W^O$ is a linear projection matrix. 
The output of each layer further goes through a feed-forward network \texttt{FFN} with a residual connection: $$O_l^r=\tilde{O}_l^r+\texttt{FFN}(\tilde{O}_l^r).$$  Layer normalization is omitted for the notational simplicity. 

\begin{figure}[ht!]
    \centering
    \includegraphics[width=.7\columnwidth]{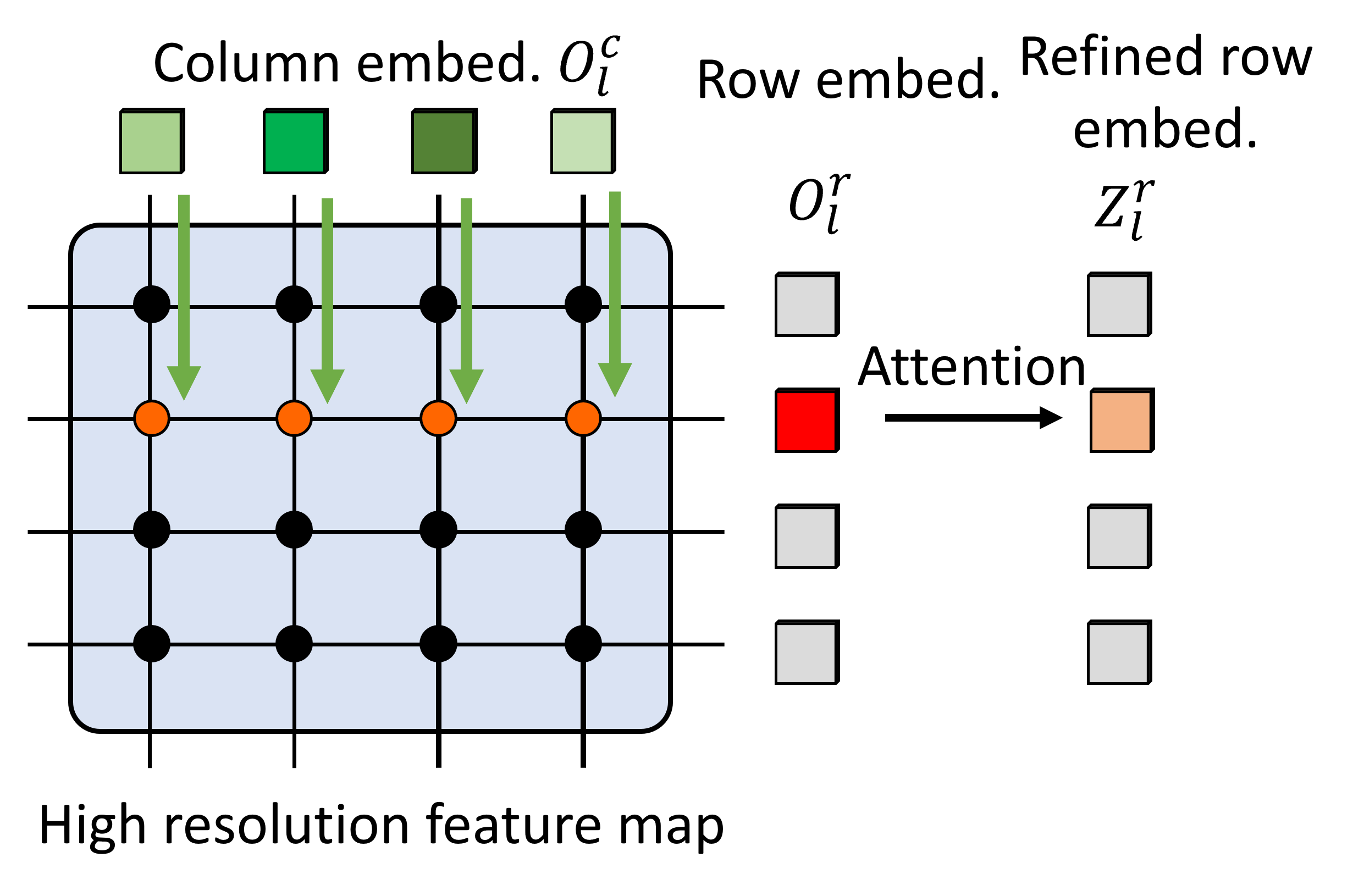}
    \caption{Illustration of row-column interactive attention, where each row(column) crosses and interacts with all columns(rows). Row-column interactive attention aims to capture these interactions to integrate information from the crossed columns (rows) to build row (column) representations.  }
    \label{fig: row_col_interact}
\end{figure}

{\noindent\bf Row-column interactive attention.} After a multi-head attention module, we also have a row-column interactive attention. As illustrated in Figure~\ref{fig: row_col_interact}, this interactive attention aims to capture the relevant information when a row (column) crosses all columns (rows) spatially. This allows the learned row (column) representation to further integrate the vertical (horizontal) contexts along the crossed columns (rows) through an interactive attention mechanism.
%This captures to reflect that rows and columns cross each other in the spatial map and 
%integrate row and column outputs through  attention. 
% \begin{align}
%     \mathcal{H}(z_q, z_k, z_v)=\texttt{SoftMax}\big(\frac{z_qW_q(z_kW_k)^T}{\sqrt{d}}\big)(z_vW_v)+z_q, 
% \end{align}
% where $W_q$, $W_k$ and $W_v$ are the linear projections.

 Formally, the row and column outputs from their interactive attentions can be obtained below,
\begin{align}
    Z_l^r=\texttt{SoftMax}\big(\frac{O_l^r{O_l^c}^T}{\sqrt{d}}\big)O_l^c+O_l^r,\\ \nonumber
    Z_l^c=\texttt{SoftMax}\big(\frac{O_l^c{O_l^r}^T}{\sqrt{d}}\big)O_l^r+O_l^c.
\end{align}
% \begin{align}
% \tilde{O}_l^r=\mathcal{H}(O_l^r, O_l^c, O_l^c),~  \tilde{O}_l^c=\mathcal{H}(O_l^c, O_l^r, O_l^r).  
% \end{align}
This module can be seen as taking the intermediate row output $O_l^r$ as the query to aggregate the relevant information from all crossing columns. For simplicity, we do not use linear projections as in multi-head attention to map $O_l^r$ and $O_l^c$ into query, key and value sequences.

%enhances the row and column outputs by considering the correlation between the two. 

{\noindent\bf Final dense feature map.} At the output end, each pixel at $(i,j)$ in the final feature map is represented as a direct combination of the outputs from the row and column transformers at the same location. Formally, it can be written as 
$$S_{ij}=Z_{L,i}^r+Z_{L,j}^c,$$
where $S_{ij}$ is the final feature vector in the dense map at $(i,j)$, and $Z_{L,i}^r$, $Z_{L,j}^c$ are the representation of row $i$ and column $j$ in the last layer, respectively. 

Note that through row-column interactive attention, each row (column) representation has already aggregated the information from all column (rows) before it is finally added to the column (row) representations to generate the dense feature output. Thus, the DFlatFormer does not wait until the final layer to re-couple rows and columns. The early interaction between rows and columns allows them to fully explore the vertical and horizontal contexts across layers while still keeping the 
DFlatFormer tractable. 
% {\small
% \begin{align*}
%     \mathcal{F}(\tilde{O}^r_{L-1},\tilde{O}^c_{L-1})=\texttt{Col\_expan}(\tilde{O}^r_{L-1})+\texttt{Row\_expan}(\tilde{O}^c_{L-1}),
% \end{align*}
% }
% where $\texttt{Row\_expan}$ and $\texttt{Col\_expan}$ represent the row and column expansion, respectively.

% \begin{figure}[ht!]
%     \centering
%     \includegraphics[width=\columnwidth]{figure/row_col_fusion2.png}
%     \caption{Illustration of row and column-wise DFlatFormer feature fusion. }
%     \label{fig: row_col_fusion}
% \end{figure}
\subsection{Efficient attentions via grouping and pooling} \label{sec:key_selection}
If all the pixels in the encoder output are taken to form keys/values, the computational complexity is $\mathcal{O}(hw(H+W))$, as there are totally $H$ rows and $W$ columns for queries and $hw$ keys/values fed into the DFlatFormer. To further reduce complexity, we propose to perform multi-head attention via grouping and (average) pooling.  Grouping divides both queries and the input feature map into several groups, where  each query can only access features within the corresponding feature group. 
On the other hand, the features in a row and a column are average-pooled to form shorter flattened sequences, further reducing the model complexity.
%from all rows/columns of the input  allow each query to access more long-range contexts.

\begin{figure}[ht!]
    \centering
    \includegraphics[width=\columnwidth]{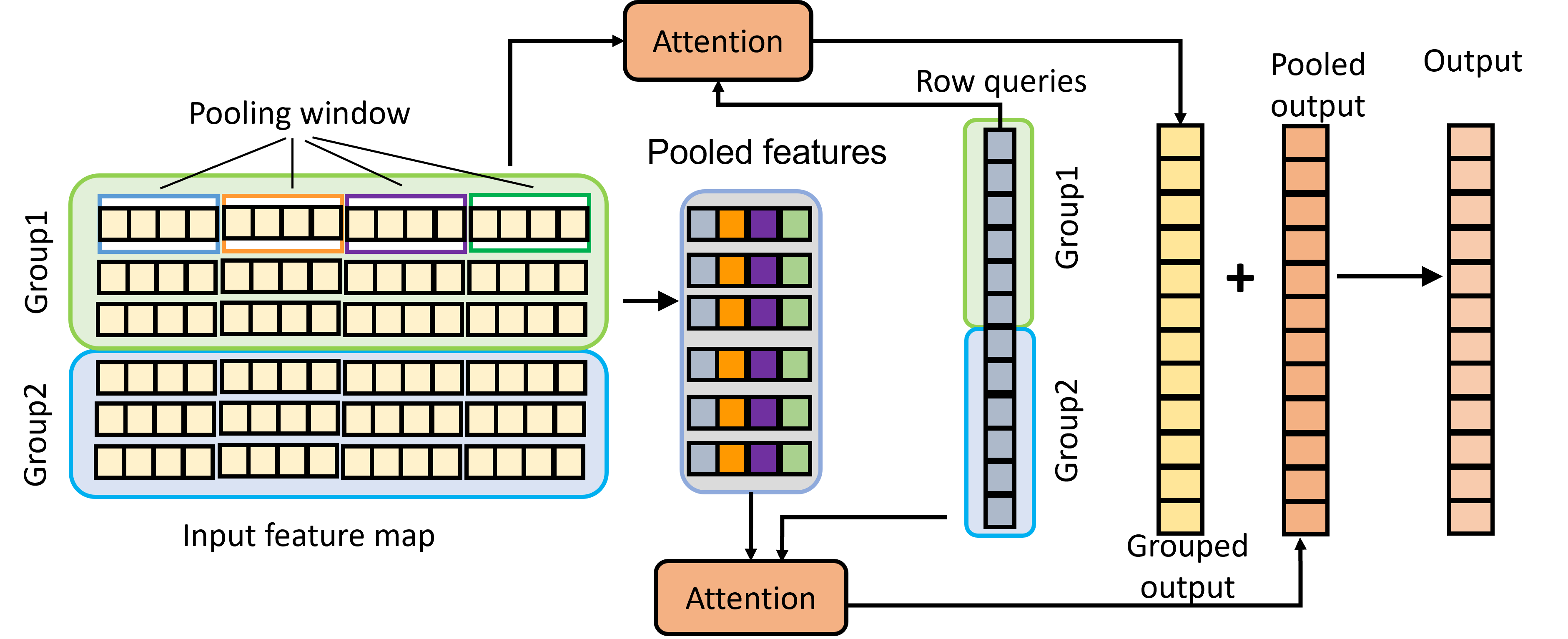}
    \caption{Attentions through grouping and pooling. }
    \label{fig:key_design}
\end{figure}

Specifically, as shown in Fig.~\ref{fig:key_design}, for a row transformer, the row queries and the row-flattened sequence are equally divided into $n_p$ groups. Multi-head attention is conducted within each group in parallel such that a row query only performs the multi-head attention on the corresponding group of the flattened sequence. On the other hand, we also average-pool the row-flattened sequence through a non-overlapping window of size $n_w$ over each row, resulting in a shorter sequence where each row query can access the whole one for performing the multi-head attention. The resultant outputs from both grouping and pooling are added to give the output. 

We note that the grouping and pooling compliment with each other -- in the grouping, each query accesses a smaller but part of the flattened sequence at its original grained level. In contrast, in the pooling, the query access the whole sequence but at a coarse level with pooled features. By combining both, the output can reach a good balance between the computational cost and the representation ability.

\subsection{Complexity analysis}
% \noindent {\bf Complexity analysis.} 
As aforementioned, with decomposed row and column queries, the overall complexity of DFlatFormer is reduced to $\mathcal{O}\big(hw(H+W)\big)$. Next let us analyze the complexity with the grouping and pooling.  Let $\beta_g=1/n_p$ be the fraction of the features within each group over the flattened sequence, and $\beta_p=1/n_w$ be the fraction of pooled features over the sequence. %We have $\beta_g=1/n_p$ and $\beta_p=1/n_w$ respectively. 
Each query only accesses a group of $\beta_g hw$ features.  There are only $\beta_p hw$ pooled features in total, and they are shared among all the queries. Hence the total number of features that a query can access is $(\beta_g+\beta_p)hw$, resulting in an overall complexity of $\mathcal{O}((\beta_g+\beta_p)hw(H+W))$. In our experiment, we typically have $\beta_g=\frac{1}{4}$ and $\beta_p=\frac{1}{4}$, hence the overall complexity is reduced to $\mathcal{O}(\frac{1}{2}hw(H+W))$ by half.

\section{Experimental results}
%\subsection{Dataset}
\noindent {\bf Datasets.} Experiments are conducted on three datasets: ADE20K~\cite{zhou2017scene}, Cityscapes~\cite{cordts2016cityscapes} and Pascal VOC~2012~\cite{everingham2010pascal}. ADE20K is a large-scale scene parsing benchmark composed of 150 classes. The number of training/validation/test images is 20k/2k/3k.  Cityscapes is an urban scene understanding dataset containing 19 classes. It has 2975/500/1525 training/validation/test images. Pascal VOC consists of 21 classes, containing 1464 images with high-quality annotations, and an extra 10582 images with coarse annotations to serve as training set. The number of images for validation/testing is 1449 and 456, respectively. 

\noindent {\bf Baseline and our models. }
With the CNN as backbone, DeepLabv3+~\cite{chen2018encoder} with ResNet-50/101 are taken as the baseline models. We deploy our DFlatFormer as the decoder upon the output of DeepLabv3+ encoder. For the ResNet-50 baseline, if low-level features are explored, we implement separate DFlatFormers to take in both high-level and low-level features to generate outputs of the same spatial size. The final feature map is given by channel-wise concatenation. We present in Appendix more details of models.

% For transformer as the backbone,  UperNet with Swin-T/Swin-S~\cite{liu2021swin} and SegFormer-B2/B4~\cite{xie2021segformer} are considered as the baseline models. For a fair comparison, we adopt Swin-T/Swin-S and MiT~\cite{xie2021segformer} as encoders and implement DFlatFormer as the corresponding decoder. Notice that Swin and MiT are both hierarchical encoders which generate multi-level features. We leverage the multi-level features by feeding the feature map in each level to a separate DFlatFormer decoder and generating outputs of the same spatial size. The channel dimension for each output depends on the input dimension and a predefined threshold $d_m$. If the channel dimension of the encoder output feature map is larger than $d_m$, a $1\times 1$ convolutional layer is use to map it to a channel  of dimension $d_m$ before feeding into the DFlatFormer. The outputs from multi-level DFlatFormers are concatenated in the channel dimension to give the target dense output.

\noindent {\bf Implementation details.}
For ADE20K and Pascal VOC datasets, we crop images into a size of $520\times 520$. For Cityscapes, the crop size is also $520\times 520$, except with MiT~\cite{xie2021segformer} backbone which uses $1024\times 1024$ crop size.
% To fairly compare with SegFormer~\cite{xie2021segformer} which utilizes MiT backbone, we crop Cityscapes images into $1024\times 1024$. 
We adopt the standard pixel-wise cross-entropy loss as the loss function. More details are provided in Appendix.

% For the DFlatFormer structure, we implement 10 layers. The channel dimension threshold $d_m$ is set to 256 if it is connected to hierarchical encoders. 
% We run each experiment on 8 Nvidia Tesla V100 cards. Polynomial learning rate scheduling is adopted with a total of 320K iterations. For ResNet-50/101 backbone, we employ SGD optimizer with a learning rate of 0.01, a momentum of 0.9 and a weight decay of $10^{-4}$. The learning rate of the decoder is 10 times that of the backbone, with a batch size of 32.  
% With Swin or MiT transformer backbones, AdamW~\cite{loshchilov2017decoupled} optimizer is used with a learning rate of $6\times 10^{-5}$, a momentum of 0.9 and a weight decay of 0.01.  The batch size is 32 for ADE20K/Pascal VOC, and 16 for Cityscapes, respectively. Following SegFormer, a model pretrained on ImageNet is taken as the encoder. For the multi-scale evaluation, we follow the standard procedure~\cite{chen2017rethinking} to scale images with multiple ratios together with horizontal flipping. Sliding window  is employed to handle varying input size during inference, where the window size matches the training crop size.

% In all experiments, we adopt efficient attentions via the grouping and pooling by default if not specified.  The encoder output feature map is divided into 4 groups, i.e, $\beta_l=\frac{1}{4}$.  For pooled features, the feature map is average pooled with $\beta_g=\frac{1}{4}$ with each row/column. Hence the total number of features that each query accesses is reduced by half.

\subsection{Main results}
We adopt the mean intersection over union (mIoU) to measure segmentation performance. To compare memory sizes and computation complexities, the number of parameters (``Param") and Giga-floating-point operations (``GFLOPs") are used. ``Size" $a$ in all tables, if not specified, denotes a crop size of $a\times a$. ``mIoU" in the tables refers to mIoU  with single-scale inference, and ``+MS" denotes mIoU with multi-scale inference and horizontal flipping.
% \begin{table*}[ht]
% \centering
% \caption{Semantic segmentation mIoU(\%) for Pascal VOC~2012 dataset with encoder DeepLabv3+ ResNet-50. Baseline is DeepLabv3+ with ResNet-50 with bilinear upsampling. }
% \begin{tabular}{llllll}
% % \begin{tabular}{p{0.25\linewidth}p{0.25\linewidth}p{0.15\linewidth}p{0.15\linewidth}p{0.15\linewidth}p{0.1\linewidth}}
% \hline
%   Scheme   & Query &Model size(M) &FLOPs(G) & Embed. dim  &miou (\%)  \\
%         \hline
% DeepLabv3+ (ResNet-50) &-- &40.94 &-- & &76.3 \\
% \hline
% Transformer &Learnable window &49.19 & &512 &76.0 \\
% Transformer &Fixed embed. &49.19 & &512 &77.9 \\
% Transformer &Learnable embed. &49.51 & &512 &77.8 \\
% Transformer(multi-level) &Learnable & & & & \\
% \hline
% \end{tabular}
% \end{table*}

% \begin{table*}[t]
% \centering
% \caption{Semantic segmentation mIoU(\%) for Pascal VOC~2012 dataset with encoder DeepLabv3+ ResNet-101. Baseline is DeepLabv3+ with ResNet-101  with bilinear upsampling. }
% \begin{tabular}{llllll}
% % \begin{tabular}{p{0.25\linewidth}p{0.25\linewidth}p{0.15\linewidth}p{0.15\linewidth}p{0.15\linewidth}p{0.1\linewidth}}
% \hline
%   Scheme   & Query &Model size(M) &FLOPs(G) & Embed. dim  &miou (\%)  \\
%         \hline
% DeepLabv3+ (ResNet-101) &-- &-- &-- & &78.5 \\
% \hline
% Transformer &Learnable window & & & & \\
% Transformer &Fixed embed. & & & & \\
% Transformer &Learnable embed. & & & & \\
% Transformer(multi-level) &Learnable & & & & \\
% \hline
% \end{tabular}
% \end{table*}

In Table~\ref{tab: dupsample_vs_dflat}  we compare the DFlatFormer with ResNet-50  baseline based on  bilinear upsampling and DUpsample~\cite{tian2019decoders}. Without low-level features, DFlatFormer gives 4.5\% performance gain over bilinear upsampling  and 3.5\% gain over DUpsample. With exploration on low-level features, DFlatFormer provides a  0.5\% additional gain in mIoU. A further comparison of DFlatFormer with PointRend~\cite{kirillov2020pointrend} is given in Appendix.  The above results suggest that DFlatFormer outperforms both bilinear upsampling and data-dependent upsampling in recovering high-resolution features for semantic segmentation. 
% Table~\ref{tab: pointrend_vs_dfat} compares the DFlatFormer with the DeepLabv3 baseline and PointRend~\cite{kirillov2020pointrend} on Cityscapes \emph{val} dataset. ResNet-101 is chosen as the backbone. It is observed that DFlatFormer outperforms the baseline by 2.1\%, and outperforms PointRend by 0.9\%. The above provides strong evidences that DFlatFormer can effectively model the long-range interactions between pixels in high-resolution feature map, providing additional gains in recovering fine-details for segmentation. 
In Table~\ref{tab:dflat_vs_deeplab_all} we compare performances between the DFlatFormer with DeepLabv3+ on three datasets: Pascal VOC 2012, Cityscapes and ADE20K. All methods adopt the ResNet-50 as backbone. DFlatFormer achieves 2.4\%, 2.0\% and 3.3\% gains over DeepLabv3+ respectively. The results suggest that DFlatFormer can well capture the global contexts and achieve universal gains across datasets. 

In the following we make more comparisons between our model and other state-of-art architectures, and show that the DFlatFormer can be used as a universal plug-in decoder for highly-accurate dense predictions for segmentation.

\begin{table}[t]
\centering
\footnotesize
\caption{Comparison of DFlatFormer with bilinear and DUpsample~\cite{tian2019decoders} on Pascal VOC \emph{val} dataset, with backbone ResNet-50. }
\vspace{-1mm}
\begin{tabular}{lcl}
% \begin{tabular}{p{0.25\linewidth}p{0.25\linewidth}p{0.15\linewidth}p{0.15\linewidth}p{0.15\linewidth}p{0.1\linewidth}}
\hline
  Scheme &w/ low-level feat. &mIoU \\
        \hline
Bilinear & &72.2 \\
DUpsample~\cite{tian2019decoders}  & &73.2 \\
DUpsample~\cite{tian2019decoders}  &\checkmark &74.2 \\
\rowcolor{LightGrey}
DFlatFormer(ours) & &76.7 \\
\rowcolor{LightGrey}
DFlatFormer(ours) &\checkmark &\textbf{77.2} \\
\hline
\end{tabular}
\label{tab: dupsample_vs_dflat}
\end{table}

% \begin{table}[t]
% \centering
% \footnotesize
% \caption{Comparison of DFlatFormer with DeepLabv3 and PointRend~\cite{kirillov2020pointrend} on Cityscapes dataset, with backhone ResNet-101. }
% \vspace{-1mm}
% \begin{tabular}{ll}
% % \begin{tabular}{p{0.25\linewidth}p{0.25\linewidth}p{0.15\linewidth}p{0.15\linewidth}p{0.15\linewidth}p{0.1\linewidth}}
% \hline
%   Method     &mIoU \\
%         \hline
% DeepLabv3-OS-16~\cite{chen2018encoder}   &77.2 \\
% PointRend~\cite{kirillov2020pointrend}  
%   &78.4 \\
% \rowcolor{LightGrey}
% DFlatFormer(ours)  &\textbf{79.3} \\
% \hline
% \end{tabular}
% \label{tab: pointrend_vs_dfat}
% \end{table}

\begin{table}[t]
\centering
\footnotesize
\caption{Semantic segmentation of DFlatFormer compared with DeepLabv3+ across three datasets, with backbone ResNet-50.  }
\vspace{-1mm}
\begin{tabular}{llll}
\hline
  Model   &Pascal VOC  &Cityscapes  &ADE20K \\
        \hline
DeepLabv3+~\cite{chen2018encoder} &76.3 &77.6 &41.5 \\
\hline
\rowcolor{LightGrey}
DFlatFormer(ours)  &\textbf{78.7}~\textcolor{blue}{(+2.4)} &\textbf{79.6}~\textcolor{blue}{(+2.0)} &\textbf{44.8}~\textcolor{blue}{(+3.3)} \\
\hline
% \end{tabularx}
\end{tabular}
 \label{tab:dflat_vs_deeplab_all}
\end{table}

\noindent {\bf Comparison With CNN backbones.}
In Table~\ref{tab:miou_cnn_ade} and Table~\ref{tab:miou_cnn_city}, we show a comprehensive comparison of DFlatFormer with other models on ADE20K and Cityscapes \emph{val} datasets, respectively. All models adopt the CNN backbones.  For ADE20K segmentation, with ResNet-50 as backbone, DFlatFormer outperforms DeepLabv3+ baseline by 3.3\% and 3.7\% for single-scale and multi-scale inference, respectively; with ResNet-101 as backbone, DFlatFormer outperforms DeepLabv3+ by 2.7\% for single-scale inference. For Cityscapes semantic segmentation, DFlatFormer with single-scale inference outperforms DeepLabv3+ by 2.0\% and 1.8\% with backbone ResNet-50 and ResNet-101, respectively. 

\noindent {\bf Comparison with Transformer backbone.}
Given transformer as backbone, Table~\ref{tab:miou_ade_transformer} and Table~\ref{tab:miou_city_transformer}  provide performance comparisons of DFlatFormer with other architectures on ADE20K \emph{val} and Cityscapes \emph{val} datasets, respectively. For ADE20K segmentation, with Swin-T as backbone, DFlatFormer has 2.6\% gain over UperNet for single-scale inference. With Swin-S as backbone, DFlatFormer surpasses UperNet by 0.6\%. On the ohter hand, the model size and GFLOPs of DFlatFormer are much smaller than the baselines. When comparing with SegFormer~\cite{xie2021segformer}, with MiT-B2 as backbone, DFlatFormer outperforms SegFormer by 0.9\%. With MiT-B4 as backbone, DFlatFormer outperforms SegFormer by 0.5\%. While the model size is slightly increased over SegFormer, our model enjoys a much smaller GFLOPs. More detailed structure comparison with SegFormer are provided in Sec.~\ref{sec:ablation}. Note that in SegFormer, \textit{AlignedResize}~\cite{xie2021segformer} is utilized which potentially provides extra gain over the normal inference. For fair comparison with others we follow the normal procedure as in MMSegmentation~\cite{mmseg2020}. The results demonstrate that DFlatFormer can further leverage the multi-level features to strength the power of hierarchical structures in dense prediction.

\begin{figure}[!t]
\center
\footnotesize
\begin{tabularx}{\textwidth}{c @{\hskip 3pt} c}
\includegraphics[ width=0.5\linewidth, keepaspectratio]{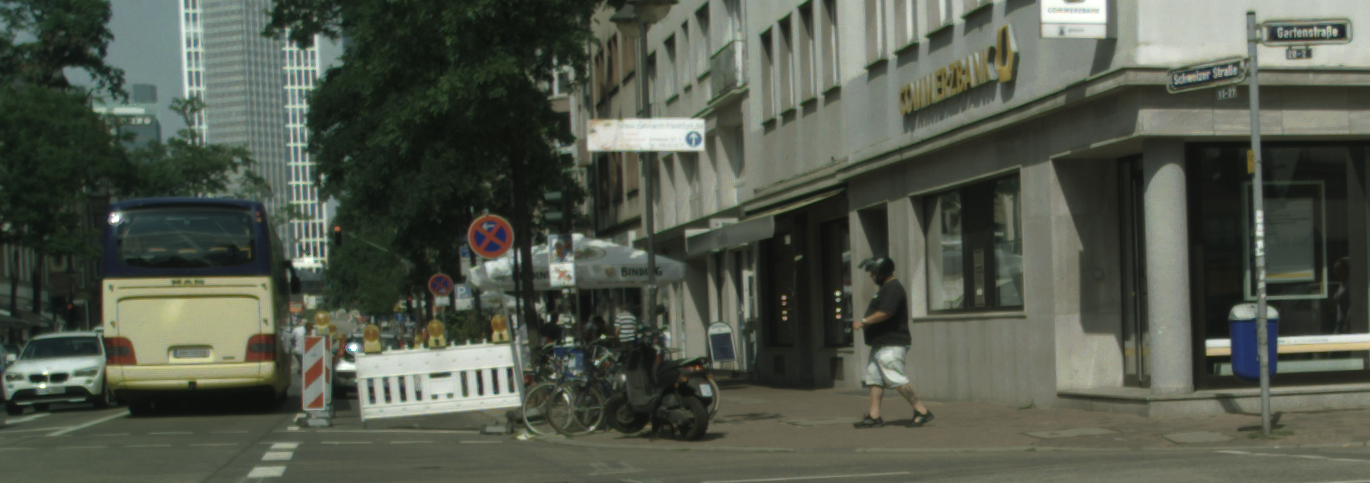}  &
\includegraphics[ width=0.5\linewidth, keepaspectratio]{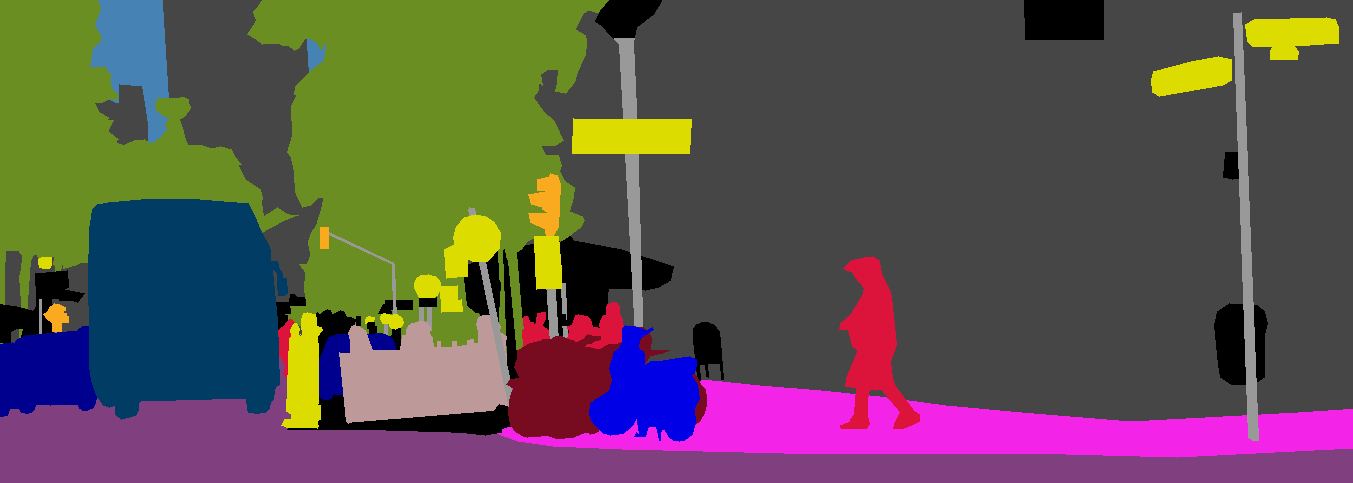} \\
\includegraphics[ width=0.5\linewidth, keepaspectratio]{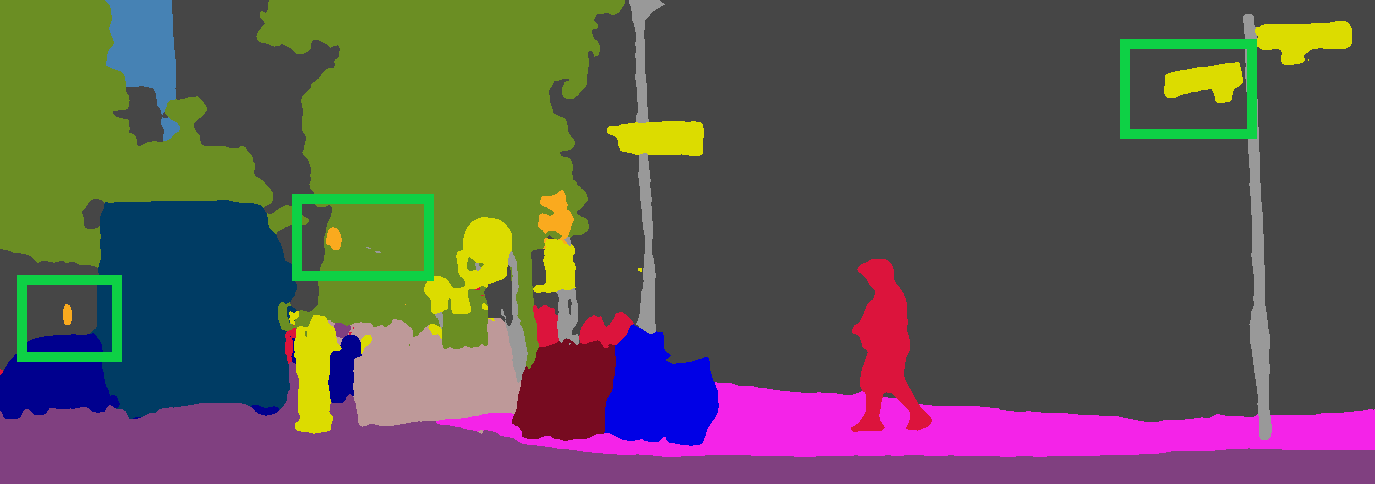}  &
\includegraphics[ width=0.5\linewidth, keepaspectratio]{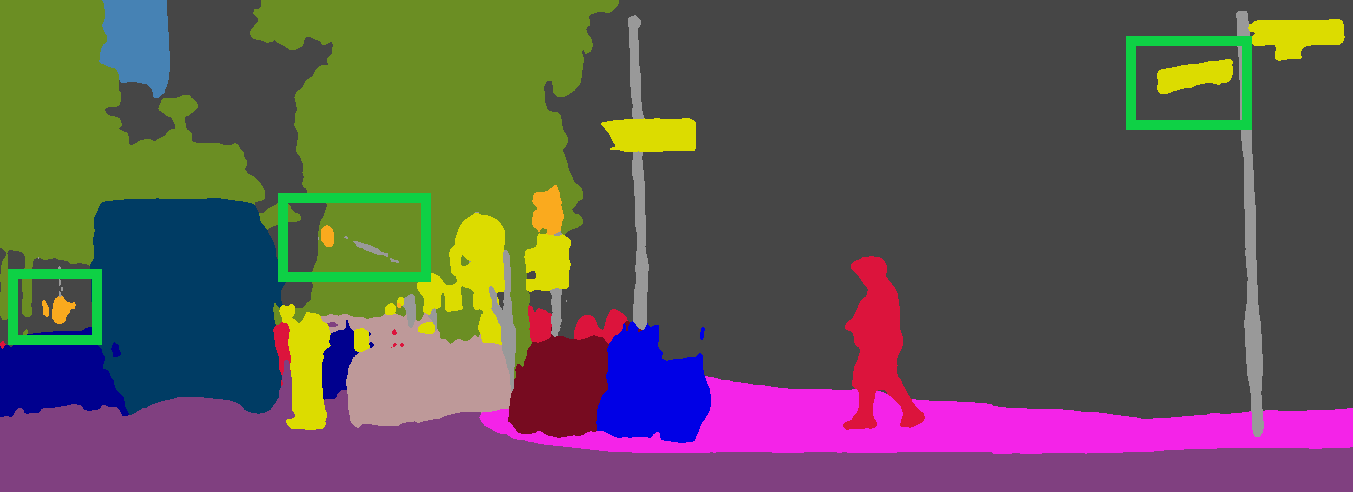} \\
\end{tabularx}
\caption{Cityscapes results for semantic segmentation. Top Left: RGB image; Top right: Ground-truth label; Bottom left: SegFormer; Bottom right: DFlatFormer. DFlatFormer provides better predictions for small objects such as traffic signs and lights.} \label{fig:cityscapes_vis}
\end{figure}

\begin{figure}[t]
\center
\footnotesize
\resizebox{\linewidth}{!}{%
\begin{tabularx}{\linewidth}{P{0pt}c @{\hskip 2pt}c@{\hskip 2pt} c }
\rotatebox{90}{RGB image} &
\includegraphics[ width=0.32\linewidth, keepaspectratio]{figure/vis/ade_vis/ADE_val_00000163_rbg.png}  &
\includegraphics[ width=0.32\linewidth, keepaspectratio]{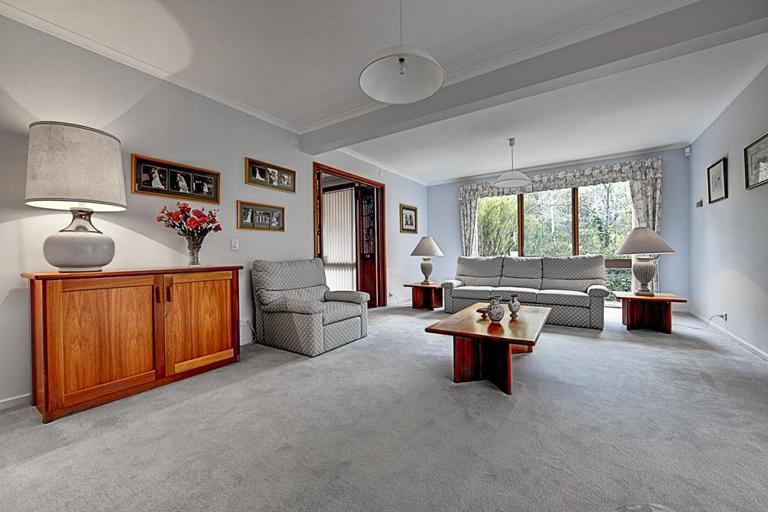}  &
\includegraphics[ width=0.30\linewidth, keepaspectratio]{figure/vis/ade_vis/ADE_val_00001171_rbg.png}  \\
\rotatebox{90}{Ground-truth}& \includegraphics[ width=0.32\linewidth, keepaspectratio]{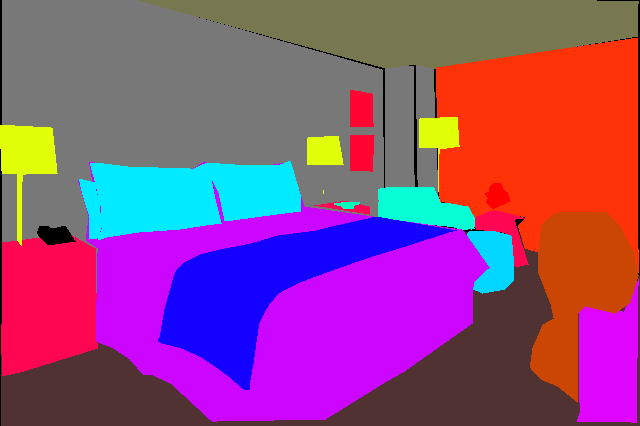} &
\includegraphics[ width=0.32\linewidth, keepaspectratio]{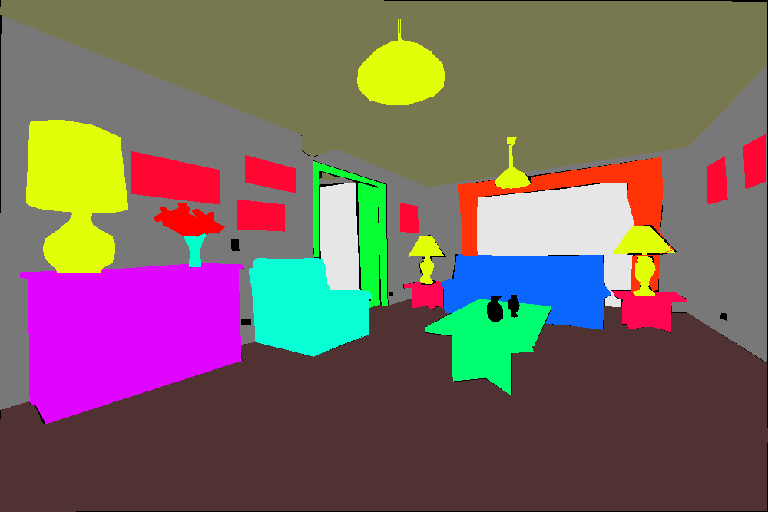} &
\includegraphics[ width=0.30\linewidth, keepaspectratio]{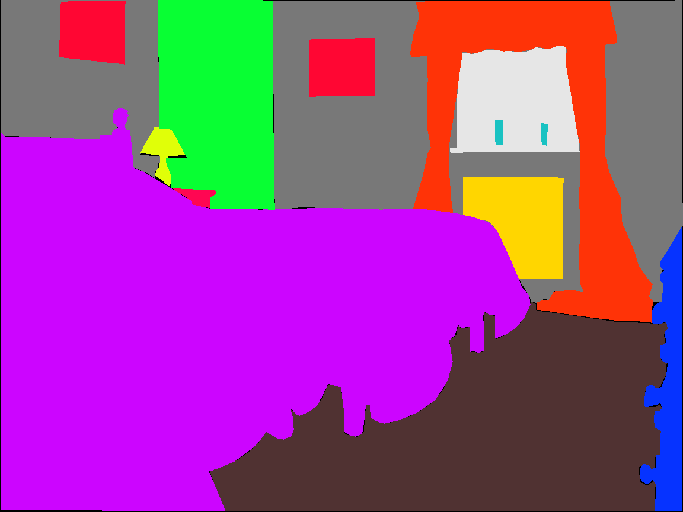} \\
\rotatebox{90}{DeepLabv3+}& \includegraphics[ width=0.32\linewidth, keepaspectratio]{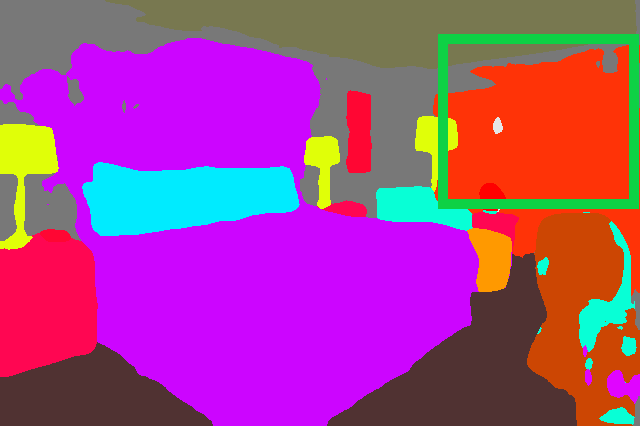} &
\includegraphics[ width=0.32\linewidth, keepaspectratio]{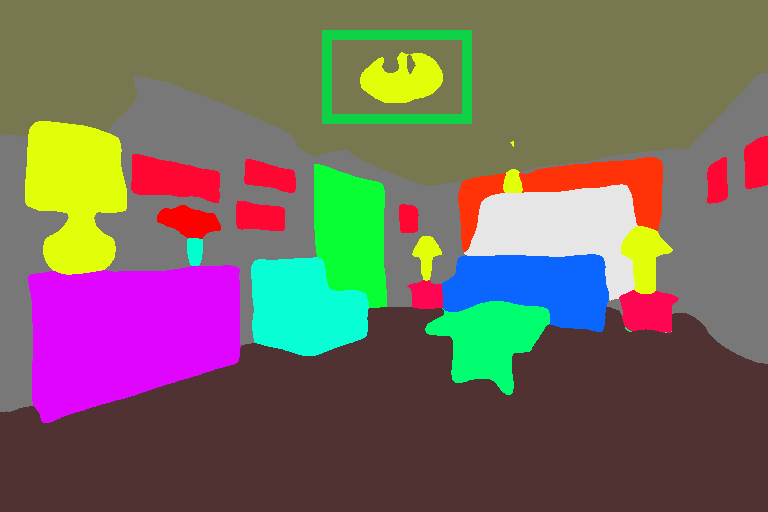}  &
\includegraphics[ width=0.28\linewidth, keepaspectratio]{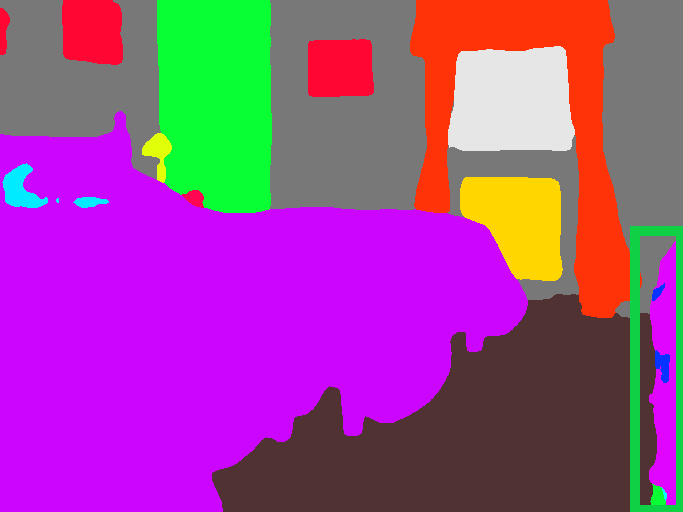} \\
\rotatebox{90}{SegFormer}& \includegraphics[ width=0.32\linewidth, keepaspectratio]{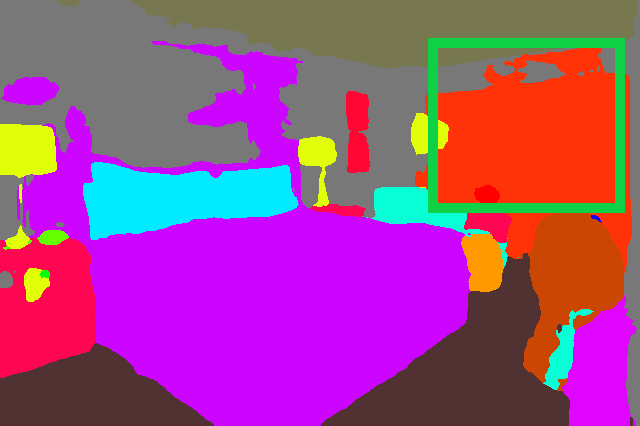} &
\includegraphics[ width=0.32\linewidth, keepaspectratio]{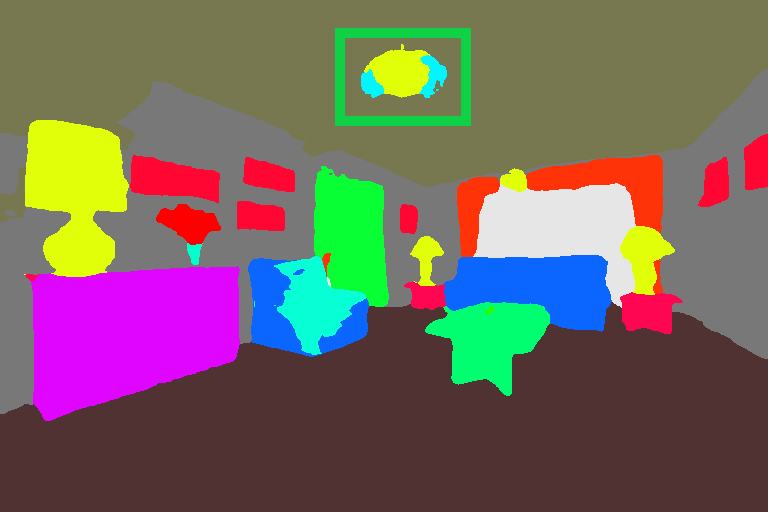} &
\includegraphics[ width=0.28\linewidth, keepaspectratio]{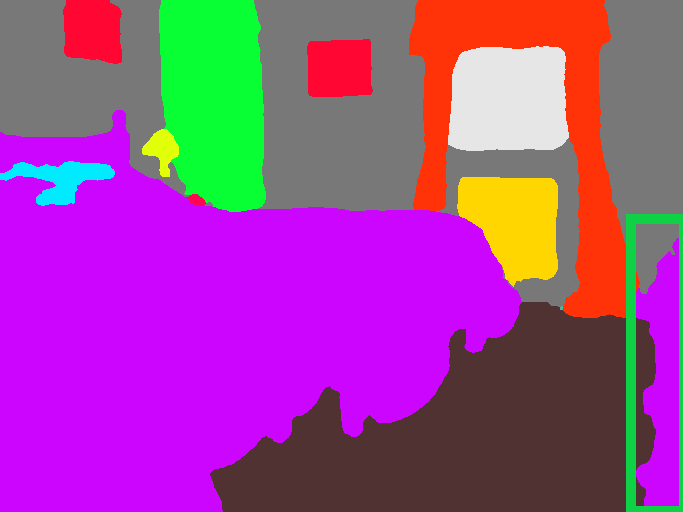} \\
\rotatebox{90}{DFlatFormer}& \includegraphics[ width=0.32\linewidth, keepaspectratio]{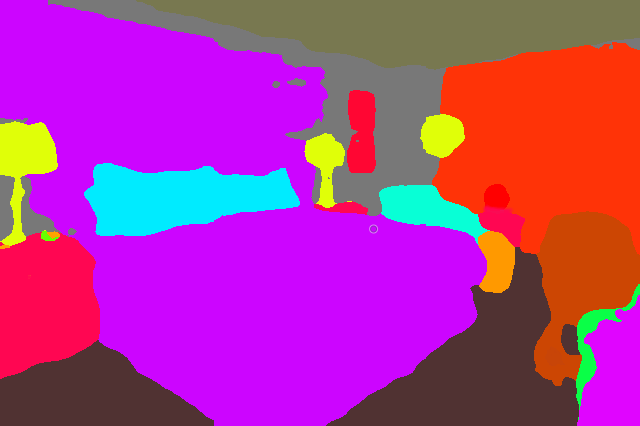}&
\includegraphics[ width=0.32\linewidth, keepaspectratio]{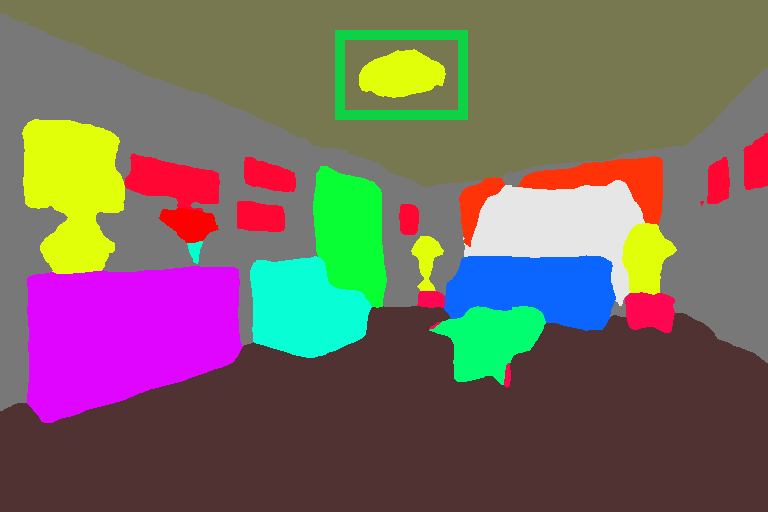} &
\includegraphics[ width=0.28\linewidth, keepaspectratio]{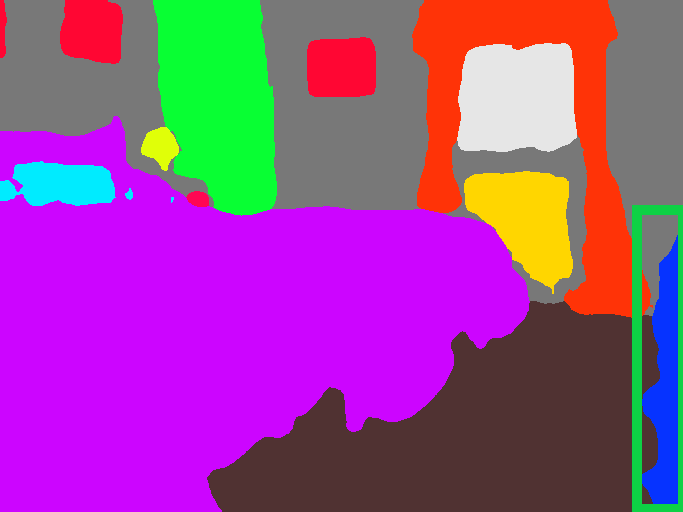} \\

% &
% \includegraphics[ width=0.32\linewidth, keepaspectratio]{figure/vis/ade_vis/ADE_val_00000163_segformer_pred.png} &
% \includegraphics[ width=0.19\linewidth, keepaspectratio]{figure/vis/ade_vis/ADE_val_00000163_dflat_pred.png}\\
% \includegraphics[ width=0.19\linewidth, keepaspectratio]{figure/vis/ade_vis/ADE_val_00001517_rbg.png}  &
% \includegraphics[ width=0.19\linewidth, keepaspectratio]{figure/vis/ade_vis/ADE_val_00001517_label.png} &
% \includegraphics[ width=0.19\linewidth, keepaspectratio]{figure/vis/ade_vis/ADE_val_00001517_deeplab_pred.png}  &
% \includegraphics[ width=0.19\linewidth, keepaspectratio]{figure/vis/ade_vis/ADE_val_00001517_segformer_pred.png} &
% \includegraphics[ width=0.19\linewidth, keepaspectratio]{figure/vis/ade_vis/ADE_val_00001517_dflat_pred.png}\\
% \includegraphics[ width=0.19\linewidth, keepaspectratio]{figure/vis/ade_vis/ADE_val_00001171_rbg.png}  &
% \includegraphics[ width=0.19\linewidth, keepaspectratio]{figure/vis/ade_vis/ADE_val_00001171_label.png} &
% \includegraphics[ width=0.19\linewidth, keepaspectratio]{figure/vis/ade_vis/ADE_val_00001171_deeplab_pred.png}  &
% \includegraphics[ width=0.19\linewidth, keepaspectratio]{figure/vis/ade_vis/ADE_val_00001171_segformer_pred.png} &
% \includegraphics[ width=0.19\linewidth, keepaspectratio]{figure/vis/ade_vis/ADE_val_00001171_dflat_pred.png}\\
% RGB input &Ground-truth &DeepLabv3+ &SegFormer &DFlatFormer \\
\end{tabularx}}
\caption{ADE20K segmentation examples for different schemes. DFlatFormer has better predictions than DeepLabv3+ and SegFormer for highlighted objects: first image: more complete prediction of curtain; second image: better shape prediction of lamp; third image: DFlatFormer accurately predicted chest of drawers, while other two failed.
 } \label{fig:ade_vis}
\end{figure}

\noindent{\bf Visualization results.}
For qualitative comparison, we visualize the segmentation results in Fig.~\ref{fig:cityscapes_vis} to compare DFlatFormer with SegFormer. It can be seen that DFlatFormer provides better predictions for small objects such as traffic signs and lights. 
%Small objects typically have higher demand for information from high-resolution features. 
In Fig.~\ref{fig:ade_vis} more comparisons are presented for DeepLabv3+ (ResNet-101), SegFormer and DFlatFormer on ADE20K. From the first two images we observe that DFlatFormer can  predict more accurate and complete boundaries for objects such as curtain and lamp.  From the last image we observe that among the three methods, only DFlatFormer can accurately predict the segmentation for chest of drawers. The results  show that DFlatFormer can effectively capture more fine-details through long-range attentions on the contexts. 

\begin{table}[t]\setlength{\tabcolsep}{2pt}
\centering
\scriptsize
\caption{Semantic segmentation performance on ADE20K \emph{val} dataset with CNN backbone.}
\vspace{-1mm}
\begin{tabular}{lllllll}

\hline
Method                        & Backbone       &Size               & Param(M) & GFLOPs & mIoU  &+MS         \\ \hline
DeepLabv3+~\cite{chen2018encoder} &ResNet-50 &512 &43.6 &176.4  &41.5 &43.2 \\
MaskFormer~\cite{cheng2021per}                    & ResNet-50 &512 &41 &-- &44.5 &46.7\\ 
\rowcolor{LightGrey}
DFlatFormer(ours) &ResNet-50 &512 &45.4 &407.4 &\textbf{44.8}\textcolor{blue}{(+3.3)} &\textbf{46.9}\textcolor{blue}{(+3.7)}\\
\hline
Auto-DeepLab-L~\cite{liu2019auto} &NAS-F48 &513 &-- &-- &44.0 &-- \\
DeepLabv3+~\cite{chen2018encoder}                    & ResNet-101       &512            & 62.7          & 255.1        & 43.2 &44.1 \\
CCNet~\cite{huang2019ccnet} & ResNet-101                  & --            &--   &--  &--  & 45.2           \\
DANet~\cite{fu2019dual} &ResNet-101 &-- &69 &1119 &45.3 &--  \\
DNL~\cite{yin2020disentangled} &ResNet-101 &-- &69 &1249 &46.0 &--  \\
ACNet~\cite{hu2019acnet} &ResNet-101 &-- &-- &-- &45.9  &-- \\
SPNet~\cite{hou2020strip} &ResNet-101 &-- &-- &-- &45.6 &--  \\
UperNet~\cite{xiao2018unified} &ResNet-101 &-- &86 &1029 &44.9 &-- \\
OCRNet~\cite{yuan2020object} &ResNet-101 &520  &56 &923 &45.3 &--  \\
MaskFormer~\cite{cheng2021per}                   & ResNet-101  &512 & 60         & --   &45.5 &47.2\\ 
OCRNet~\cite{yuan2020object} &HRNet-w48 &520 &71 &664 &45.7 &--  \\

\rowcolor{LightGrey}
DFlatFormer(ours) &ResNet-101 &512 &68.3 &512.6 &\textbf{45.9}\textcolor{blue}{(+2.7)} &\textbf{47.2}\textcolor{blue}{(+3.1)} \\
\hline
\end{tabular}
\label{tab:miou_cnn_ade}
\end{table}

\begin{table}[ht]\setlength{\tabcolsep}{2pt}
\centering
\scriptsize
\caption{Semantic segmentation on Cityscapes \emph{val} dataset with CNN backbone. $\dag$ denotes a crop size of 512x1024.}
\begin{tabular}{lllllll}

\hline
Method                       & Backbone      &Size                &Param(M) & GFLOPs & mIoU &+MS          \\ \hline
DeepLabv3+~\cite{chen2018encoder} &ResNet-50 &512 &43.6 &176.4 &77.6 &78.4\\
DUpsample~\cite{tian2019decoders} &ResNet-50 &-- &-- &-- &79.1 &--\\
\rowcolor{LightGrey}
DFlatFormer(ours) &ResNet-50 &512 &45.4 &407.4 &\textbf{79.6}\textcolor{blue}{(+2.0)} &\textbf{80.9}\textcolor{blue}{(+2.5)} \\
\hline
Auto-DeepLab-L~\cite{liu2019auto} &NAS-F48 &769 &44.4 &695.0 &80.3 &--\\
DeepLabv3+~\cite{chen2018encoder}                    & ResNet-101    &512              & 62.7          & 255.1           & 78.8 &79.3 \\
CCNet~\cite{huang2019ccnet} & ResNet-101    & 769             & 68.9            & 224.8      &-- & 80.5           \\
GCNet~\cite{cao2019gcnet} &ResNet-101 &-- &-- &-- &-- &78.1 \\
DANet~\cite{fu2019dual} &ResNet-101 &769 &-- &-- &-- &81.5 \\
OCRNet~\cite{yuan2020object} &ResNet-101 & 769 &-- &-- &-- &81.8 \\
% Axial-DeepLab-S~\cite{wang2020axial} &Axial-ResNet-50 &-- &12.1 & 220.8 &80.5 &-- \\
% Axial-DeepLab-L~\cite{wang2020axial} &Axial-ResNet-50 &-- &44.9 & 687.4 &81.0 &-- \\
% Dynamic Routing~\cite{li2020learning} &Dynamic-L33-PSP &-- &--  &270.0 &80.7 &-- \\
MaskFormer~\cite{cheng2021per}                    & ResNet-101 &512\dag & 60        &--     & 78.5     &--          \\

\rowcolor{LightGrey}
DFlatFormer(ours) &ResNet-101 &512 &68.3 &512.6 &\textbf{80.6}\textcolor{blue}{(+1.8)} &\textbf{81.9}\textcolor{blue}{(+2.6)} \\
\hline
\end{tabular}
\label{tab:miou_cnn_city}
\end{table}

\begin{table}[ht]\setlength{\tabcolsep}{1.5pt}
\centering
\scriptsize
% \scriptsize
\caption{Semantic segmentation performance on ADE20K \emph{val}  dataset, with transformer backbone. $\dag$: model pretrained on ImageNet-22K. $\ast$: \textit{AlignedResize} used in inference.}
\vspace{-1mm}
\begin{tabular}{lllllll}
\hline
Method                        & Backbone       &Size               & Param(M) & GFLOPs & mIoU  &+MS         \\ 
\hline
$\text{SETR}^{\dag}$~\cite{zheng2021rethinking}                         & ViT-L        &--   & 308          &--   &48.6        & 50.3             \\
PVT+Trans2Seg &PVT-S~\cite{wang2021pyramid} &-- &32.1 &31.6 &42.6 &--\\

Semantic FPN~\cite{kirillov2019panoptic} &PVT-M~\cite{wang2021pyramid} &512 &48.0 &219.0 &41.6 &--\\
Semantic FPN &P2T-S~\cite{wu2021p2t} &512 &26.8 &41.7  &44.4 &--\\
Semantic FPN &P2T-B~\cite{wu2021p2t} &512 &39.9 &57.5  &46.2 &--\\
Semantic FPN &Swin-S~\cite{liu2021swin} &512 &53.2 &270.4 &45.2 &-- \\
Semantic FPN &CrossFormer-S~\cite{wang2021crossformer} &512 &34.3 &220.7 &46.0 &-- \\
% Semantic FPN &CrossFormer-B~\cite{wang2021crossformer} &512x512 &55.6 &331.0 &47.7 &-- \\
% Semantic FPN &CrossFormer-L~\cite{wang2021crossformer} &512x512 &95.4 &497.0 &48.7 &--\\
% UperNet &CSWin-T~\cite{dong2021cswin} &512x512 &59.9 &959 &49.3 &50.4 \\
% UperNet &CSWin-S~\cite{dong2021cswin} &512x512 &64.6 &1027 &50.0 &50.8 \\
% % UperNet &Twins-SVT-S~\cite{chu2021twins}  &512x512 &54.4 &228 &46.2 &47.1 \\
% % UperNet &Twins-SVT-B~\cite{chu2021twins} &512x512  &88.5 &261 &47.7 &48.9\\
% % UperNet &Twins-SVT-L~\cite{chu2021twins}  &512x512 &133 &297 &48.8 &50.2 \\
% UperNet &Focal-T~\cite{yang2021focal} &640x640 &62 &998 &45.8 &47.0 \\
% UperNet &Focal-S~\cite{yang2021focal} &640x640 &85 &1130 &48.0 &50.0 \\
UperNet & CrossFormer-S~\cite{wang2021crossformer} &512 &62.3 &979.5 &47.6 &48.4 \\
% UperNet &CrossFormer-B~\cite{wang2021crossformer} &512x512 &83.6 &1089.7 &49.7 &50.6 \\
% UperNet &CrossFormer-L~\cite{wang2021crossformer} &512x512 &125.5 &1257.8 &50.4 &51.4 \\

UperNet &Swin-T~\cite{liu2021swin}  &512 &60 &945 &44.5 &46.1 \\

MaskFormer~\cite{cheng2021per} &Swin-T &512 &42 &-- &46.7 &\textbf{48.8} \\
\rowcolor{LightGrey}
DFlatFormer(ours) &Swin-T  &512 &53.1 &523.2 &\textbf{47.1}\textcolor{blue}{(+2.6)} &48.4\textcolor{blue}{(+2.3)} \\
UperNet &Swin-S~\cite{liu2021swin} &512 &81.3 &1038 &47.6 &49.3 \\
MaskFormer~\cite{cheng2021per} &Swin-S &512 &63 &-- &\textbf{49.8} &\textbf{51.0} \\
\rowcolor{LightGrey}
DFlatFormer(ours) &Swin-S &512 &72.8 &719.6 &48.3\textcolor{blue}{(+0.6)} &49.8\textcolor{blue}{(+0.5)} \\
\hline
Semantic FPN &CrossFormer-B~\cite{wang2021crossformer} &512 &55.6 &331.0 &47.7 &-- \\
Semantic FPN &CrossFormer-L~\cite{wang2021crossformer} &512 &95.4 &497.0 &48.7 &--\\
UperNet &CSWin-T~\cite{dong2021cswin} &512 &59.9 &959 &49.3 &50.4 \\
UperNet &CSWin-S~\cite{dong2021cswin} &512 &64.6 &1027 &50.0 &50.8 \\
% UperNet &Twins-SVT-S~\cite{chu2021twins}  &512x512 &54.4 &228 &46.2 &47.1 \\
UperNet &Twins-SVT-B~\cite{chu2021twins} &512  &88.5 &261 &47.7 &48.9\\
UperNet &Twins-SVT-L~\cite{chu2021twins}  &512 &133 &297 &48.8 &50.2 \\
UperNet &Focal-T~\cite{yang2021focal} &640 &62 &998 &45.8 &47.0 \\
UperNet &Focal-S~\cite{yang2021focal} &640 &85 &1130 &48.0 &50.0 \\
SegFormer~\cite{xie2021segformer} &MiT-B2 &512 &27.5 &62.4 &$46.5^\ast$ &47.5 \\
\rowcolor{LightGrey}
DFlatFormer(ours)  &MiT-B2 &512 &30.5 &48.4 &\textbf{47.4}\textcolor{blue}{(+0.9)} &\textbf{48.2}\textcolor{blue}{(+0.7)} \\ 
% SegFormer~\cite{xie2021segformer} &MiT-B3 &512x512 &47.3 &79.0 &49.4 &50.0 \\
SegFormer~\cite{xie2021segformer} &MiT-B4 &512 &64.1 &95.7 &$50.3^\ast$ &51.1 \\
% SegFormer~\cite{xie2021segformer} &MiT-B5 &512x512 &84.7 &183.3 &51.0 &51.8 \\

\rowcolor{LightGrey}
DFlatFormer(ours)  &MiT-B4 &512 &72.7 &78.4 &\textbf{50.8}\textcolor{blue}{(+0.5)} &\textbf{51.7}\textcolor{blue}{(+0.6)} \\
\hline
\end{tabular}
\label{tab:miou_ade_transformer}
\end{table}

\begin{table}[ht]\setlength{\tabcolsep}{2pt}
\scriptsize
\centering
\caption{Semantic segmentation on Cityscapes \emph{val} dataset with transformer backbone. $\ddag$: crop size of $512\times 1024$.}
\vspace{-1mm}
\begin{tabular}{lllllll}

\hline 
Method                       & Backbone      &Size                & Param(M) & GFLOPs & mIoU &+MS          \\ \hline
Semantic FPN &PVT-T~\cite{wang2021pyramid} &$\text{1024}^\ddag$ &17.0 &-- &71.7 &--\\
Semantic FPN &P2T-T~\cite{wu2021p2t} &$\text{1024}^\ddag$ &14.9 &-- &77.3 &--\\
Semantic FPN &PVT-S~\cite{wang2021pyramid}&$\text{1024}^\ddag$ &28.2 &-- &75.5 &--\\
Semantic FPN &P2T-S~\cite{wu2021p2t} &$\text{1024}^\ddag$ &26.8 &-- &78.6 &--\\
Semantic FPN &PVT-M~\cite{wang2021pyramid} &$\text{1024}^\ddag$ &48.0 &-- &76.6 &--\\
Semantic FPN &P2T-B~\cite{wu2021p2t} &$\text{1024}^\ddag$ &39.9 &-- &79.7 &--\\
SETR~\cite{zheng2021rethinking}                          & ViT    &--       & 97.6          &--           & 79.5        &--      \\
SETR~\cite{zheng2021rethinking} &ViT-L &-- &318.3 &-- &82.2 &--\\
SegFormer~\cite{xie2021segformer} &MiT-B2 &1024 &27.5 &717.1 &81.0 &82.2 \\
\rowcolor{LightGrey}
DFlatFormer(ours)  &MiT-B2 &1024 &36.2 &384.2 &\textbf{81.9}\textcolor{blue}{(+0.9)} &\textbf{82.8}\textcolor{blue}{(+0.6)} \\
SegFormer~\cite{xie2021segformer}                     & MiT-B4  &1024   & 64.1          &1240.6   & 82.3 &83.8              \\
% SegFormer~\cite{xie2021segformer} &MiT-B5 &1024x1024 &84.7 &1460.4 &82.4 &84.0 \\

% Trans4Trans                   & PVTv2-B3                      & 49.55M         & 94.25G     & 81.54              \\

\rowcolor{LightGrey}
DFlatFormer(ours)  &MiT-B4  &1024 &72.8 &612.6 &\textbf{82.8}\textcolor{blue}{(+0.5)} &\textbf{84.5}\textcolor{blue}{(+0.7)} \\
% \rowcolor{LightGrey}
% DFlatFormer(ours) &Swin-T &1024x1024 & 0&0 &0 &0\\
\hline
\end{tabular}
\label{tab:miou_city_transformer}
\end{table}

% \begin{figure}[t]
% \center
% \begin{tabularx}{\textwidth}{c c}
% \includegraphics[ width=0.5\linewidth, keepaspectratio]{figure/vis/frankfurt_000001_067474_rbg.png}  &
% \includegraphics[ width=0.5\linewidth, keepaspectratio]{figure/vis/frankfurt_000001_067474_label - Copy.png} \\
% \includegraphics[ width=0.5\linewidth, keepaspectratio]{figure/vis/out_segform_only_opacity0frankfurt_000001_067474_pred - Copy.png}  &
% \includegraphics[ width=0.5\linewidth, keepaspectratio]{figure/vis/frankfurt_000001_067474_pred - Copy.png} \\
% \end{tabularx}
% \caption{Cityscapes results for semantic segmentation. Top Left: RGB image; Top right: Ground-truth label; Bottom left: SegFormer; Bottom right: DFlatFormer.} \label{fig:cityscapes_vis_full}
% \end{figure}

\subsection{Ablation studies}\label{sec:ablation}
\noindent {\bf Attentions through grouping and pooling.} We conduct ablation study on attentions through grouping and pooling in Table~\ref{tab:miou_group_pool_interact} for ADE20K. Performance with grouping and pooling is slightly better the one without them. We hypothesize that grouped features enable queries to interact more closely with spatially correlated features from neighboring rows and columns, while pooled features efficiently aggregate 
 information from local pixels before served to model global  contextual details through the DFlatFormer.

\noindent{\bf Row-column interactive attentions. } We compare the performance of DFlatFormer with and without row-column interactions in Table~\ref{tab:miou_group_pool_interact} on ADE20K \emph{val} dataset. Backbones ResNet-50 and MiT-B2 are employed.  It is observed that with interactive attentions, performance increases by 1.5\% and 1.3\% with backbone ResNet-50 and MiT-B2, respectively. Results demonstrate the importance of earlier interactions between rows and columns before they are eventually combined at the output end.

% \begin{table}[t]
% \centering
% \footnotesize
% \caption{DFlatFormer  performance with row-column interactive attentions.}
% \vspace{-1mm}
% \begin{tabular}{lcl}
% % \begin{tabular}{p{0.25\linewidth}p{0.25\linewidth}p{0.15\linewidth}p{0.15\linewidth}p{0.15\linewidth}p{0.1\linewidth}}
% \hline
%   Backbone &Row-column interactions  &mIoU  \\
%         \hline
% \multirow{2}{*}{ResNet-50}   & &43.3  \\
%  & \checkmark  &44.8 \\
% \hline
% \multirow{2}{*}{MiT-B2}   & &46.1 \\
%  &\checkmark  &47.4 \\
% \hline
% \end{tabular}
% \label{tab:miou_interact}
% \end{table}

\begin{table}[t]
\centering
\footnotesize
\caption{DFlatFormer performance with grouping and pooling attentions, and with row-column interactive attentions.}
\vspace{-1mm}
\begin{tabular}{P{.14\columnwidth}|P{.18\columnwidth}|P{.1\columnwidth}||P{.18\columnwidth}|P{.1\columnwidth} }
% \begin{tabular}{p{0.25\linewidth}p{0.25\linewidth}p{0.15\linewidth}p{0.15\linewidth}p{0.15\linewidth}p{0.1\linewidth}}
\hline
  Backbone &Grouping and pooling attentions  &mIoU &Row-column interactions &mIoU \\
        \hline
\multirow{2}{*}{ResNet-50}   & &44.5 & &43.3 \\
 & \checkmark  &44.8 & \checkmark  &44.8 \\
\hline
\multirow{2}{*}{MiT-B2} &  &47.3 & &46.1\\
 &\checkmark  &47.4  &\checkmark  &47.4\\
\hline
\end{tabular}
\label{tab:miou_group_pool_interact}
\end{table}

\noindent {\bf Comparison with SegFormer decoder.}
SegFormer employs a light-weighted MLP decoder with output of all-level features concatenated. Although its model size is smaller than DFlatFormer, its GFLOPs is much larger than ours. Detailed comparisons of model size and GFLOPs are presented in Table~\ref{tab:segformer_vs_dflat}. We notice that an embedding dimension of 768 is employed for each hierarchical level in SegFormer to achieve decent performance. In DFlatFormer, we only set the embedding dimension up to 256. %where $c_i$ is the $i$-th level input channel dimension. 
As there are four levels of output features, after the channel-wise concatenation,  SegFormer has the total number of channels much larger than ours, leading to heavier computational cost as well as memory footprint although it has a smaller number of model parameters. 

\begin{table}[t]
\centering
\footnotesize
\caption{Comparison of DFlatFormer and SegFormer with backbone MiT-B2.}
\begin{tabular}{l|c|c|c}
\hlineB{2}
\multirow{2}{*}{Method}&Performance&\multicolumn{2}{c}{Decoder size and complexity}\\
        \cline{2-4}
&mIoU &Param(M) &GFLOPs\\
\hline
SegFormer &46.5  &3.3 &42.1 \\
DFlatFormer(ours) &47.4 &6.1 &27.1\\
\hline
\end{tabular}
\label{tab:segformer_vs_dflat}
\end{table}
 
 \noindent{\bf Ablations on other factors.} The comparisons of other factors are presented in Appendix, including positional codes and number of decoder layers, etc.

\section{Conclusion}
We propose a dual-flattening transformer to learn dense feature maps for high-resolution semantic segmentation. Given an input feature map of size $h\times w$ and an output target of size $H\times W$, the proposed model merely incurs a complexity of $\mathcal{O}\big(hw(H+W)\big)$  by employing decomposed row and column queries to capture global attentions. Input feature maps are row-wise and column-wise flattened to maintain their row and column structures, and hence the decomposed queries can retrieve immediate contexts from neighboring rows and columns. To recover the spatial details of high resolution, we introduce row-column interactive attentions at their crossing points. We further present efficient attentions via grouping and pooling in the feature space to further reduce the complexity. Experiments have demonstrated superior performances of the proposed DFlatFormer over others with both CNN and Transformer backbones.

% \clearpage
%%%%%%%%% REFERENCES
{\small
\bibliographystyle{ieee_fullname}
\bibliography{ref_trans}

\begin{thebibliography}{10}\itemsep=-1pt

\bibitem{bello2019attention}
Irwan Bello, Barret Zoph, Ashish Vaswani, Jonathon Shlens, and Quoc~V Le.
\newblock Attention augmented convolutional networks.
\newblock In {\em Proceedings of the IEEE/CVF international conference on
  computer vision}, pages 3286--3295, 2019.

\bibitem{beltagy2020longformer}
Iz Beltagy, Matthew~E Peters, and Arman Cohan.
\newblock Longformer: The long-document transformer.
\newblock {\em arXiv preprint arXiv:2004.05150}, 2020.

\bibitem{cao2019gcnet}
Yue Cao, Jiarui Xu, Stephen Lin, Fangyun Wei, and Han Hu.
\newblock Gcnet: Non-local networks meet squeeze-excitation networks and
  beyond.
\newblock In {\em Proceedings of the IEEE/CVF International Conference on
  Computer Vision Workshops}, pages 0--0, 2019.

\bibitem{carion2020end}
Nicolas Carion, Francisco Massa, Gabriel Synnaeve, Nicolas Usunier, Alexander
  Kirillov, and Sergey Zagoruyko.
\newblock End-to-end object detection with transformers.
\newblock In {\em European Conference on Computer Vision}, pages 213--229.
  Springer, 2020.

\bibitem{chen2018encoder}
Liang-Chieh Chen, Yukun Zhu, George Papandreou, Florian Schroff, and Hartwig
  Adam.
\newblock Encoder-decoder with atrous separable convolution for semantic image
  segmentation.
\newblock In {\em Proceedings of the European conference on computer vision
  (ECCV)}, pages 801--818, 2018.

\bibitem{cheng2021per}
Bowen Cheng, Alexander~G Schwing, and Alexander Kirillov.
\newblock Per-pixel classification is not all you need for semantic
  segmentation.
\newblock {\em arXiv preprint arXiv:2107.06278}, 2021.

\bibitem{child2019generating}
Rewon Child, Scott Gray, Alec Radford, and Ilya Sutskever.
\newblock Generating long sequences with sparse transformers.
\newblock {\em arXiv preprint arXiv:1904.10509}, 2019.

\bibitem{chollet2017xception}
Fran{\c{c}}ois Chollet.
\newblock Xception: Deep learning with depthwise separable convolutions.
\newblock In {\em Proceedings of the IEEE conference on computer vision and
  pattern recognition}, pages 1251--1258, 2017.

\bibitem{chu2021twins}
Xiangxiang Chu, Zhi Tian, Yuqing Wang, Bo Zhang, Haibing Ren, Xiaolin Wei,
  Huaxia Xia, and Chunhua Shen.
\newblock Twins: Revisiting the design of spatial attention in vision
  transformers.
\newblock {\em arXiv preprint arXiv:2104.13840}, 1(2):3, 2021.

\bibitem{mmseg2020}
MMSegmentation Contributors.
\newblock {MMSegmentation}: Openmmlab semantic segmentation toolbox and
  benchmark.
\newblock \url{https://github.com/open-mmlab/mmsegmentation}, 2020.

\bibitem{cordts2016cityscapes}
Marius Cordts, Mohamed Omran, Sebastian Ramos, Timo Rehfeld, Markus Enzweiler,
  Rodrigo Benenson, Uwe Franke, Stefan Roth, and Bernt Schiele.
\newblock The cityscapes dataset for semantic urban scene understanding.
\newblock In {\em Proceedings of the IEEE conference on computer vision and
  pattern recognition}, pages 3213--3223, 2016.

\bibitem{dong2021cswin}
Xiaoyi Dong, Jianmin Bao, Dongdong Chen, Weiming Zhang, Nenghai Yu, Lu Yuan,
  Dong Chen, and Baining Guo.
\newblock Cswin transformer: A general vision transformer backbone with
  cross-shaped windows.
\newblock {\em arXiv preprint arXiv:2107.00652}, 2021.

\bibitem{dosovitskiy2020image}
Alexey Dosovitskiy, Lucas Beyer, Alexander Kolesnikov, Dirk Weissenborn,
  Xiaohua Zhai, Thomas Unterthiner, Mostafa Dehghani, Matthias Minderer, Georg
  Heigold, Sylvain Gelly, et~al.
\newblock An image is worth 16x16 words: Transformers for image recognition at
  scale.
\newblock In {\em International Conference on Learning Representations}, 2020.

\bibitem{everingham2010pascal}
Mark Everingham, Luc Van~Gool, Christopher~KI Williams, John Winn, and Andrew
  Zisserman.
\newblock The pascal visual object classes (voc) challenge.
\newblock {\em International journal of computer vision}, 88(2):303--338, 2010.

\bibitem{fu2019dual}
Jun Fu, Jing Liu, Haijie Tian, Yong Li, Yongjun Bao, Zhiwei Fang, and Hanqing
  Lu.
\newblock Dual attention network for scene segmentation.
\newblock In {\em Proceedings of the IEEE/CVF Conference on Computer Vision and
  Pattern Recognition}, pages 3146--3154, 2019.

\bibitem{hou2020strip}
Qibin Hou, Li Zhang, Ming-Ming Cheng, and Jiashi Feng.
\newblock Strip pooling: Rethinking spatial pooling for scene parsing.
\newblock In {\em Proceedings of the IEEE/CVF Conference on Computer Vision and
  Pattern Recognition}, pages 4003--4012, 2020.

\bibitem{hu2019acnet}
Xinxin Hu, Kailun Yang, Lei Fei, and Kaiwei Wang.
\newblock Acnet: Attention based network to exploit complementary features for
  rgbd semantic segmentation.
\newblock In {\em 2019 IEEE International Conference on Image Processing
  (ICIP)}, pages 1440--1444. IEEE, 2019.

\bibitem{huang2019ccnet}
Zilong Huang, Xinggang Wang, Lichao Huang, Chang Huang, Yunchao Wei, and Wenyu
  Liu.
\newblock {CCNet}: Criss-cross attention for semantic segmentation.
\newblock In {\em Proceedings of the IEEE/CVF International Conference on
  Computer Vision}, pages 603--612, 2019.

\bibitem{kirillov2019panoptic}
Alexander Kirillov, Ross Girshick, Kaiming He, and Piotr Doll{\'a}r.
\newblock Panoptic feature pyramid networks.
\newblock In {\em Proceedings of the IEEE/CVF Conference on Computer Vision and
  Pattern Recognition}, pages 6399--6408, 2019.

\bibitem{kirillov2020pointrend}
Alexander Kirillov, Yuxin Wu, Kaiming He, and Ross Girshick.
\newblock Pointrend: Image segmentation as rendering.
\newblock In {\em Proceedings of the IEEE/CVF conference on computer vision and
  pattern recognition}, pages 9799--9808, 2020.

\bibitem{kitaev2019reformer}
Nikita Kitaev, Lukasz Kaiser, and Anselm Levskaya.
\newblock Reformer: The efficient transformer.
\newblock In {\em International Conference on Learning Representations}, 2019.

\bibitem{lin2017feature}
Tsung-Yi Lin, Piotr Doll{\'a}r, Ross Girshick, Kaiming He, Bharath Hariharan,
  and Serge Belongie.
\newblock Feature pyramid networks for object detection.
\newblock In {\em Proceedings of the IEEE conference on computer vision and
  pattern recognition}, pages 2117--2125, 2017.

\bibitem{liu2019auto}
Chenxi Liu, Liang-Chieh Chen, Florian Schroff, Hartwig Adam, Wei Hua, Alan~L
  Yuille, and Li Fei-Fei.
\newblock Auto-deeplab: Hierarchical neural architecture search for semantic
  image segmentation.
\newblock In {\em Proceedings of the IEEE/CVF Conference on Computer Vision and
  Pattern Recognition}, pages 82--92, 2019.

\bibitem{liu2021swin}
Ze Liu, Yutong Lin, Yue Cao, Han Hu, Yixuan Wei, Zheng Zhang, Stephen Lin, and
  Baining Guo.
\newblock Swin transformer: Hierarchical vision transformer using shifted
  windows.
\newblock {\em arXiv preprint arXiv:2103.14030}, 2021.

\bibitem{long2015fully}
Jonathan Long, Evan Shelhamer, and Trevor Darrell.
\newblock Fully convolutional networks for semantic segmentation.
\newblock In {\em Proceedings of the IEEE conference on computer vision and
  pattern recognition}, pages 3431--3440, 2015.

\bibitem{shaw2018self}
Peter Shaw, Jakob Uszkoreit, and Ashish Vaswani.
\newblock Self-attention with relative position representations.
\newblock {\em arXiv preprint arXiv:1803.02155}, 2018.

\bibitem{tao2020hierarchical}
Andrew Tao, Karan Sapra, and Bryan Catanzaro.
\newblock Hierarchical multi-scale attention for semantic segmentation.
\newblock {\em arXiv preprint arXiv:2005.10821}, 2020.

\bibitem{tian2019decoders}
Zhi Tian, Tong He, Chunhua Shen, and Youliang Yan.
\newblock Decoders matter for semantic segmentation: Data-dependent decoding
  enables flexible feature aggregation.
\newblock In {\em Proceedings of the IEEE/CVF Conference on Computer Vision and
  Pattern Recognition}, pages 3126--3135, 2019.

\bibitem{vaswani2017attention}
Ashish Vaswani, Noam Shazeer, Niki Parmar, Jakob Uszkoreit, Llion Jones,
  Aidan~N Gomez, {\L}ukasz Kaiser, and Illia Polosukhin.
\newblock Attention is all you need.
\newblock In {\em Advances in neural information processing systems}, pages
  5998--6008, 2017.

\bibitem{wang2021max}
Huiyu Wang, Yukun Zhu, Hartwig Adam, Alan Yuille, and Liang-Chieh Chen.
\newblock Max-deeplab: End-to-end panoptic segmentation with mask transformers.
\newblock In {\em Proceedings of the IEEE/CVF Conference on Computer Vision and
  Pattern Recognition}, pages 5463--5474, 2021.

\bibitem{wang2020axial}
Huiyu Wang, Yukun Zhu, Bradley Green, Hartwig Adam, Alan Yuille, and
  Liang-Chieh Chen.
\newblock Axial-deeplab: Stand-alone axial-attention for panoptic segmentation.
\newblock In {\em European Conference on Computer Vision}, pages 108--126.
  Springer, 2020.

\bibitem{wang2020linformer}
Sinong Wang, Belinda~Z Li, Madian Khabsa, Han Fang, and Hao Ma.
\newblock Linformer: Self-attention with linear complexity.
\newblock {\em arXiv preprint arXiv:2006.04768}, 2020.

\bibitem{wang2021pyramid}
Wenhai Wang, Enze Xie, Xiang Li, Deng-Ping Fan, Kaitao Song, Ding Liang, Tong
  Lu, Ping Luo, and Ling Shao.
\newblock Pyramid vision transformer: A versatile backbone for dense prediction
  without convolutions.
\newblock {\em arXiv preprint arXiv:2102.12122}, 2021.

\bibitem{wang2021crossformer}
Wenxiao Wang, Lu Yao, Long Chen, Deng Cai, Xiaofei He, and Wei Liu.
\newblock Crossformer: A versatile vision transformer based on cross-scale
  attention.
\newblock {\em arXiv preprint arXiv:2108.00154}, 2021.

\bibitem{wu2021p2t}
Yu-Huan Wu, Yun Liu, Xin Zhan, and Ming-Ming Cheng.
\newblock {P2T}: Pyramid pooling transformer for scene understanding.
\newblock {\em arXiv preprint arXiv:2106.12011}, 2021.

\bibitem{xiao2018unified}
Tete Xiao, Yingcheng Liu, Bolei Zhou, Yuning Jiang, and Jian Sun.
\newblock Unified perceptual parsing for scene understanding.
\newblock In {\em Proceedings of the European Conference on Computer Vision
  (ECCV)}, pages 418--434, 2018.

\bibitem{xie2021segformer}
Enze Xie, Wenhai Wang, Zhiding Yu, Anima Anandkumar, Jose~M Alvarez, and Ping
  Luo.
\newblock Segformer: Simple and efficient design for semantic segmentation with
  transformers.
\newblock {\em arXiv preprint arXiv:2105.15203}, 2021.

\bibitem{yang2021focal}
Jianwei Yang, Chunyuan Li, Pengchuan Zhang, Xiyang Dai, Bin Xiao, Lu Yuan, and
  Jianfeng Gao.
\newblock Focal self-attention for local-global interactions in vision
  transformers.
\newblock {\em arXiv preprint arXiv:2107.00641}, 2021.

\bibitem{yin2020disentangled}
Minghao Yin, Zhuliang Yao, Yue Cao, Xiu Li, Zheng Zhang, Stephen Lin, and Han
  Hu.
\newblock Disentangled non-local neural networks.
\newblock In {\em European Conference on Computer Vision}, pages 191--207.
  Springer, 2020.

\bibitem{yuan2020object}
Yuhui Yuan, Xilin Chen, and Jingdong Wang.
\newblock Object-contextual representations for semantic segmentation.
\newblock In {\em Computer Vision--ECCV 2020: 16th European Conference,
  Glasgow, UK, August 23--28, 2020, Proceedings, Part VI 16}, pages 173--190.
  Springer, 2020.

\bibitem{zheng2021rethinking}
Sixiao Zheng, Jiachen Lu, Hengshuang Zhao, Xiatian Zhu, Zekun Luo, Yabiao Wang,
  Yanwei Fu, Jianfeng Feng, Tao Xiang, Philip~HS Torr, et~al.
\newblock Rethinking semantic segmentation from a sequence-to-sequence
  perspective with transformers.
\newblock In {\em Proceedings of the IEEE/CVF Conference on Computer Vision and
  Pattern Recognition}, pages 6881--6890, 2021.

\bibitem{zhou2017scene}
Bolei Zhou, Hang Zhao, Xavier Puig, Sanja Fidler, Adela Barriuso, and Antonio
  Torralba.
\newblock Scene parsing through {ADE20K} dataset.
\newblock In {\em Proceedings of the IEEE conference on computer vision and
  pattern recognition}, pages 633--641, 2017.

\end{thebibliography}
}

\end{document}